%% file: main.tex
\documentclass[letterpaper, 10 pt, journal, twoside]{IEEEtran}
\usepackage{amsmath,amsfonts}
\usepackage{algorithmic}
\usepackage{algorithm}
\usepackage{array}
\usepackage[caption=false,font=normalsize,labelfont=sf,textfont=sf]{subfig}
\usepackage{textcomp}
\usepackage{stfloats}
\usepackage{url}
\usepackage{verbatim}
\usepackage{graphicx}
\usepackage{cite}
\usepackage{placeins}

\usepackage{amssymb}
\usepackage{amstext}
\usepackage{amsmath}
\usepackage{multicol, multirow}
\usepackage{dsfont}
\usepackage{pifont}
\usepackage[dvipsnames]{xcolor}
\usepackage{overpic}
\usepackage{tikz}
\usepackage{soul}
\usepackage{url}
\usepackage{listings}
\usepackage{lipsum}
\usepackage{placeins}
\usepackage{booktabs,hyphenat} 
\usepackage{cite,hyperref}

\usepackage{adjustbox}

\usepackage[resetlabels,labeled]{multibib}
\newcites{R}{References}
\usepackage[switch]{lineno}

\newcommand{\fmname}[0]{\mbox{FMF-SLAM}}
\usepackage{soul}
\usepackage{todonotes}
\usepackage{bm}

\begin{document}



\bibliographystyleR{IEEEtran}
\ifCLASSOPTIONcaptionsoff
  \newpage
\fi
\setcounter{figure}{0}
\setcounter{table}{0}

\title{Multimodal Fusion SLAM with Fourier Attention}

 
\author{Youjie Zhou$^1$, Guofeng Mei$^2$, Yiming Wang$^2$, Yi Wan$^1$, Fabio Poiesi$^2$

\thanks{$^{1}$Youjie Zhou and Yi Wan are with the School of Mechanical Engineering, Shandong University, Jinan 250100, China. (e-mail: 202020511@mail.sdu.edu.cn; wanyi@sdu.edu.cn).}%
\thanks{$^{2}$Guofeng Mei, Yiming Wang, and Fabio Poiesi are with Fondazione Bruno Kessler, 38123 Trento, Italy (e-mail: gmei@fbk.eu; ywang@fbk.eu; poiesi@fbk.eu).}

}

\markboth{Accepted in IEEE RAL}
{Zhou \MakeLowercase{\textit{et al.}}: Multimodal Fusion SLAM with Fourier Attention}

\maketitle

\begin{abstract}
Visual SLAM is particularly challenging in environments affected by noise, varying lighting conditions, and darkness.
Learning-based optical flow algorithms can leverage multiple modalities to address these challenges, but traditional optical flow-based visual SLAM approaches often require significant computational resources.
To overcome this limitation, we propose FMF-SLAM, an efficient multimodal fusion SLAM method that utilizes fast Fourier transform (FFT) to enhance the algorithm efficiency.
Specifically, we introduce a novel Fourier-based self-attention and cross-attention mechanism to extract features from RGB and depth signals.
We further enhance the interaction of multimodal features by incorporating multi-scale knowledge distillation across modalities.
We also demonstrate the practical feasibility of FMF-SLAM in real-world scenarios with real time performance by integrating it with a security robot by fusing with a global positioning module GNSS-RTK and global Bundle Adjustment.
Our approach is validated using video sequences from TUM, TartanAir, and our real-world datasets, showcasing state-of-the-art performance under noisy, varying lighting, and dark conditions.
Our code and datasets are available at \url{https://github.com/youjie-zhou/FMF-SLAM.git}.
\end{abstract}


\input{sections/introduction}

\input{sections/related_work.tex}
\input{sections/approach.tex}
\input{sections/exp_new}

\input{sections/conclusions.tex}


\IEEEpubidadjcol

\bibliographystyle{IEEEtran}
\bibliography{refenew}

\end{document}

%% file: sections/introduction.tex
\section{Introduction}\label{sec:introduciton}

With the advancement of robotics and intelligent sensing technologies, autonomous robots are increasingly playing crucial roles in areas such as industrial automation \cite{industry}, agriculture \cite{agriculture}, and disaster response \cite{emergency_response}.
To achieve autonomy, robots must be able to navigate through unseen environments, with localization being a critical component of this process \cite{challengs}.
Visual Simultaneous Localization and Mapping (SLAM) is a key technology for navigation~\cite{ORB-SLAM2, DSO}, but it presents significant challenges when sensor noise, varying lighting conditions, and dark environments affect the visual signal~\cite{challengs}.

Multimodal-based optical flow algorithms, as standalone methods, have shown promising potential in addressing such challenges by leveraging attention mechanisms to extract features from multiple modalities~\cite{Fusionraft}.
The attention mechanism enables the model to dynamically focus on the most informative features across modalities, resulting in improved flow estimation~\cite{benefit_attention}. However, naively integrating state-of-the-art multimodal methods addressing optical flow estimation~\cite{attentive} into visual SLAM systems can incur significant computation cost. This is mainly due to the numerous dot-product operations when the model dynamically weighs the importance of various inputs.
Moreover, features from each modality are often independently extracted prior to fusion with disparities in feature magnitudes across modalities, causing inaccurate computation of the contribution of each modality~\cite{Fusionraft}.

In this work, we propose {\fmname}, a novel multimodal SLAM system to address visually-challenging environments with efficient and effective multimodal feature fusion, that enables real-world deployment on robotic platforms. 
{\fmname} builds on top of a state-of-the-art learning-based SLAM method, \textit{i.e.} DROID-SLAM~\cite{DROID_SLAM}. 
Our proposal {\fmname} advances DROID-SLAM in three key ways.
First, we introduce a novel attention mechanism based on a two-branch encoder, where we leverage Fast Fourier Transform (FFT) \cite{fouriertransform} to reduce the computation overhead inherited within the self-attention and cross-attention modules during multimodal feature extraction. 
Second, inspired by recent works that learn informative features through cross-modality distillation~\cite{crossdistillation}, we propose a multi-scale distillation method within the two-branch feature encoder to enhances the interaction and aggregation across modalities.
Finally, we integrate {\fmname} with the global positioning module GNSS-RTK \cite{GNSS-RTK} and global Bundle Adjustment (BA) \cite{ORB-SLAM2} and deploy it on a real robot. 

We demonstrate that the system can successfully perform SLAM in large outdoor environments under challenging visual conditions.
We evaluate {\fmname} on video sequences from TUM \cite{TUM}, TartanAir \cite{TartanAir}, and our own datasets. 
Our dataset was recorded under standard, light-changing, and dark conditions in indoor environments.
Our evaluation provides a comprehensive validation of {\fmname}'s localization performance in various challenging scenarios, in comparison to other SLAM systems, demonstrating state-of-the-art performance and real-time robotic deployment.

In summary, our contributions are:
\begin{itemize}
    \item We propose the first learning-based multimodal SLAM system {\fmname} with efficient and effective multimodal feature extraction and aggregation, scoring the best performance on standard and visually challenging scenes.
    \item We introduce a novel light-weight attention mechanism leveraging Fast Fourier transform to efficiently extract informative multimodal features.
    \item We propose a novel multi-scale knowledge distillation technique to align multimodal features, effectively enhancing cross-modal interaction and fusion.
    \item We deploy {\fmname} on a real robotic platform in real time, demonstrating its efficiency and robust performance in both indoor and outdoor environments. 
\end{itemize}

%% file: sections/related_work.tex
\section{Related work}\label{sec:related_work}

\subsection{Classical SLAM methods}\label{sec:related_work:Classical_methods}
\vspace{-1mm}

Classical visual SLAM primarily utilizes observations from monocular, stereo, or RGB-D images, and their approaches are commonly
categorized as direct direct \cite{DSO} and indirect \cite{ORB-SLAM2,ORB-SLAM3} methods.
Indirect methods often start with front-end modules for feature extraction and matching: they first extract features using algorithms such as SURF \cite{SURF} or ORB \cite{ORB}, which are then matched to establish correspondences across frames. 
This allows the camera pose to be estimated with back-end modules by minimizing the re-projection error via Bundle Adjustment.
In contrast, direct methods utilize the intensity values of all image pixels to estimate camera motion and 3D structure. 
These methods focus on minimizing the photometric error, which is a measure of how well the estimated camera motion aligns with the observed pixel intensities.

However, both direct and indirect methods have limitations when it comes to extracting reliable correspondences in visually challenging environments characterized by noise, varying light, and darkness in the scenes. 
Instead, our proposed method is composed with a learning-based front-end module for efficient multimodal feature extraction and fusion to overcome situations where visual information alone is limited.

\vspace{-3mm}
\subsection{Deep learning-based SLAM methods}\label{sec:related_work:learning-based}
%
Learning-based methods are emerging in the application of SLAM. There exist studies that leverage deep learning techniques to address sub-problems within the SLAM systems, for instance, front-end modules including feature extraction \cite{Superpoint} and matching \cite{Superglue}. 
For example, SuperPoint~\cite{Superpoint} automates the feature extraction process employing a deep neural network that simultaneously detects points of interest and describes their features. 
SuperGlue \cite{Superglue} utilizes a graph neural network to dynamically match a set of characteristics across images, improving the precision and reliability of feature matching in challenging conditions.
There also exist efforts in developing learning-based methods to address SLAM in an end-to-end fashion. 
DeepVO~\cite{Deepvo} estimates camera poses from consecutive frames with an end-to-end framework, while DeepV2D~\cite{Deepv2d} alternates between depth prediction and camera pose estimation, without relying on bundle adjustment. 
DeepFactors~\cite{Deepfactors} is the first complete deep SLAM system that performs joint optimization of pose and depth estimation.
Subsequent work~\cite{Tartanvo} further improves the model generalisation capability across datasets with an up-to-scale loss function leveraging camera intrinsic parameters.

Yet, generalization remains a major challenge for end-to-end localization and mapping, in particular in complex environments. 
Differently, for a better generalization capability, our proposed method combines learning-based front-end modules for multimodal feature extraction and fusion, with optimization-based back-end for pose estimation.

\vspace{-3mm}
\subsection{Learning-based front-ends with optimization back-ends}\label{sec:related_work:combine}

The combination of learning-based front-ends and classical optimization back-ends presents a novel approach that enhances generalization across diverse environments.
DVSO \cite{DVSO} formulates visual odometry as a geometric optimization problem, integrating photo consistency constraints alongside deep monocular depth prediction into the optimization process.
D3VO \cite{D3VO} estimates pose and depth using a neural network front-end, which is tightly integrated with the back-end non-linear optimization of DSO \cite{DSO}.
DROID-SLAM \cite{DROID_SLAM} estimates camera poses by predicting dense flow, exhibiting good generalization in diverse environments.
DPVO \cite{DPVO} estimates pose using an efficient optimization approach that predicts flow for corresponding patches.
Previous methods focused on extracting features or estimating pose and depth directly from RGB information in the front-end.

To the best of our knowledge, current works only rely on RGB inputs, which limits their application in visually challenging environments, such as dark or light-changing conditions.
To address this challenge, we propose a multimodal SLAM system that efficiently estimates dense flow by extracting features from both RGB and depth information. Our proposal enables robust estimation even when RGB information is unreliable.

%% file: sections/approach.tex
\section{Our approach}\label{sec:method}

\begin{figure*}[t]
\centering
\small
\includegraphics[width=1.0\textwidth]{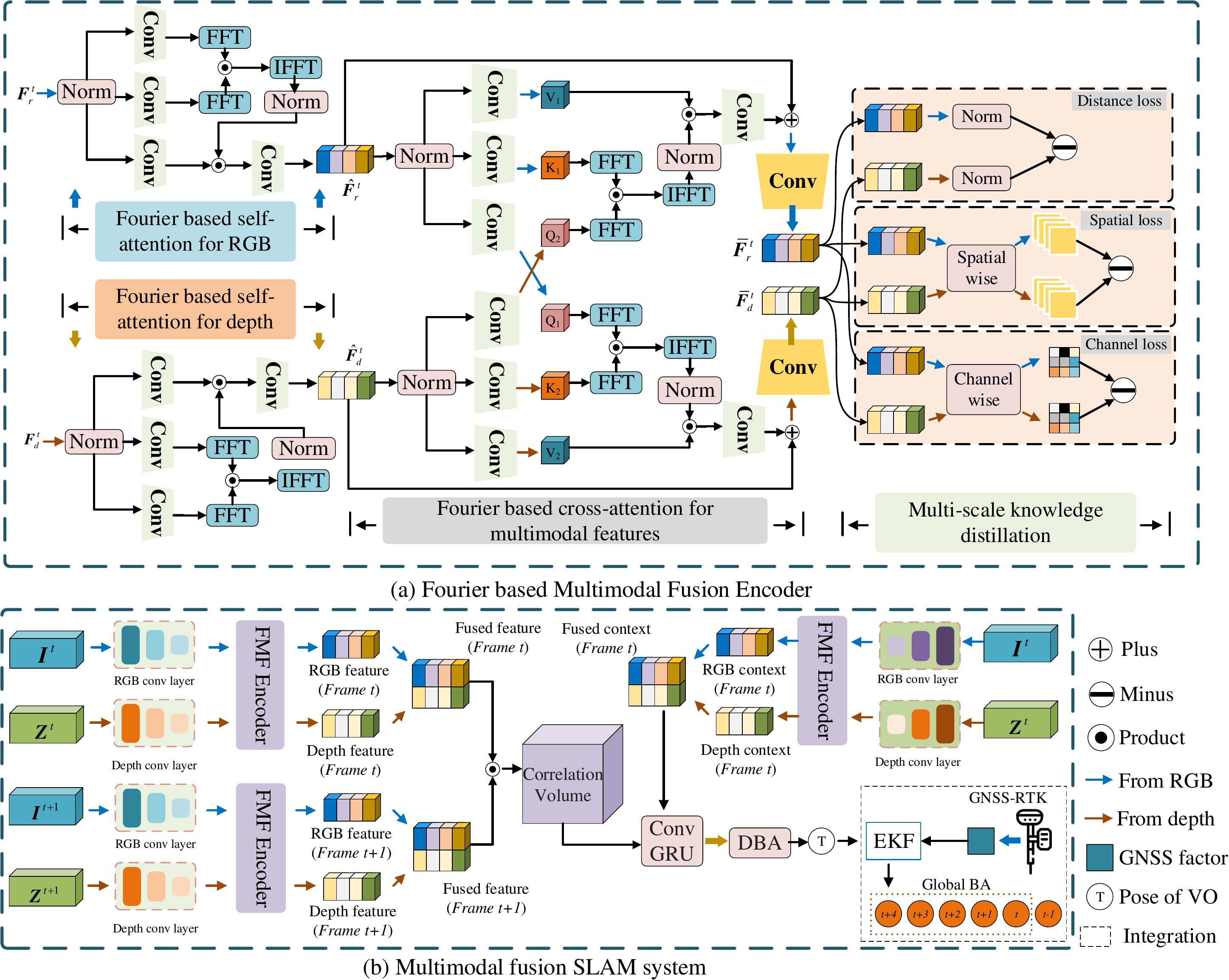}
\vspace{-3mm}
\caption{Block diagram of \fmname. (a) Encoder: Inputs are $\bm{F}^t_r$ and $\bm{F}^t_d$ (RGB and depth features) processed through Fourier-based self-attention, cross-attention, and multi-scale knowledge distillation. (b) \fmname{} architecture: Two consecutive RGB-D frames feed into the convolutional layer and encoder, while the first RGB-D frame goes to the context encoder. Dense Bundle Adjustment (DBA) refines flow estimates into pose and depth updates.}
\label{fig:method:mmf}
\vspace{-4mm}
\end{figure*}

\fmname{} leverages multimodal fusion of RGB and depth information to enhance localization and mapping in challenging environments. 
Its architecture incorporates an attention-based mechanism comprising both Fourier-based self-attention and cross-attention. 
Moreover, multi-scale knowledge distillation is integrated into the supervision process to facilitate knowledge transfer between modalities. 
For robust real-world deployment, \fmname{} is integrated within a robot using GNSS-RTK and global Bundle Adjustment. 
Fig.~\ref{fig:method:mmf}(a) depicts the architecture of our Fourier-based multimodal fusion encoder (FMF), while Fig.~\ref{fig:method:mmf}(b) illustrates the integration of \fmname{} into the robotic system.

\subsection{Fourier based multimodal feature fusion encoder}\label{sec:methods:fusion_encoder}

Given a pair of consecutive RGB-D frames $(P^t,~P^{t+1})$ at time $t$ as input, and each frame $P^t = \{I^t, Z^t\}$ is composed of a RGB image $I^t$ and a depth image $Z^t$. We first obtain low-level features $\bm{{F}}^t_r\in \mathbb{R}^{W \times H \times D}$ and $\bm{{F}}^t_d\in \mathbb{R}^{W \times H \times D}$ from each modality with convolutional blocks, where we use the subscript ${r}$ to represent the RGB branch and ${d}$ for the depth branch (Fig.~\ref{fig:method:mmf}(a)). 
${D}$ is the feature dimension and ${W \times H}$ is the resolution of the features.
\begin{equation}
\bm{F}_r^t = {\rm{Conv}}({I^t}),\bm{F}_d^t = {\rm{Conv(}}{{\rm{Z}}^t}{\rm{)}},
\end{equation}
where $\bm{F}_r^t$ and $\bm{F}_d^t$ present the feature from RGB and depth encoder, respectively. 

Attention mechanisms computationally costly due to numerous dot-product operations \cite{attentive}, which can hinder their application in visual SLAM on robotic platforms. 
We propose a Fourier-based attention mechanism that estimates attention by using element-wise product operations instead of matrix multiplication, thus reducing computational cost.

\noindent\textbf{Fourier based self-attention.}
Given the input features of RGB and depth branches with the shape ${W \times H \times D}$ , we first obtain ${\bm{Q}_{\bm{F}_i^t}}$, ${\bm{K}_{\bm{F}_i^t}}$, and ${\bm{V}_{\bm{F}_i^t}}$ with a convolutional block for RGB and depth branches, respectively. 
Then, we utilize Fast Fourier Transform (FFT) to handle ${\bm{Q}_{\bm{F}_i^t}}$ and ${\bm{K}_{\bm{F}_i^t}}$ via
\begin{equation}
\bm{A_i} = IFFT(\bm{{F}}^t_i)
 = IFFT(FFT({{\bm{Q}_{\bm{F}_i^t}}}) \overline {FFT({{\bm{K}_{\bm{F}_i^t}}})} ),
\end{equation}
where $i\in\{r,d\}$, $FFT( \cdot )$ denotes Fast Fourier Transform \cite{fouriertransform}, $IFFT( \cdot )$ denotes the inverse FFT. 
$\overline {FFT( \cdot )}$ denotes the conjugate transpose operation. 
In the frequency domain, the conjugate transpose is crucial to accurately compute the correlation between two complex-valued features. 
Also, tensors include both real and imaginary parts. 
Based on the conjugate symmetry in $FFT( \cdot )$, we only store half of the tensor data by utilizing $\overline {FFT( \cdot )}$, which prevents an increase in memory consumption\cite{1998}.
%
We compute self-attention features as
\begin{equation}
\begin{aligned}
{\bm{V}_{i}} = {\cal N}(\bm{A}_i){{\bm{V}_{\bm{F}_i^t}}},
\end{aligned}
\vspace{-1mm}
\end{equation}
where ${\cal N}( \cdot )$ denotes a layer norm that is used to normalize. 
Lastly, we generate the output feature of self-attention as
\begin{equation}
\begin{aligned}
\bm{\hat{F}}^t_i = \bm{{F}}^t_i + {\rm{Con}}{{\rm{v}}_{1 \times 1}}({\bm{V}_{i}}).
\end{aligned}
\end{equation}
Conv1$\times$1(·) denotes a 1$\times$1 pixel convolution.

\noindent\textbf{Fourier based cross-attention.}
We  highlight informative features across multimodal information by Fourier based cross-attention (Fig.~\ref{fig:method:mmf}(a)).
%
Given the input features computed from self-attention, we obtain $\bm{\hat{Q}}_{\bm{F}r^t}$ and $\bm{\hat{K}}{\bm{F}_r^t}$, and ${\bm{\hat{V}}_{\bm{F}_r^t}}$ for the RGB branch, ${\bm{\hat{Q}}_{\bm{F}_d^t}}$, ${\bm{\hat{K}}_{\bm{F}_d^t}}$, and ${\bm{\hat{V}}_{\bm{F}_d^t}}$ for the depth branch, respectively.
\begin{equation}
\begin{aligned}
{\bm{V}_{r}} = {\cal N}(IFFT(FFT({{\bm{\hat{Q}}_{\bm{F}_d^t}}}) \overline {FFT({{\bm{\hat{K}}_{\bm{F}_r^t}}})} )){{\bm{\hat{V}}_{\bm{F}_r^t}}},
\end{aligned}
\vspace{-1mm}
\end{equation}
where ${\bm{V}_{r}}$ denotes the attention map for RGB information, ${\cal N}( \cdot )$ denotes the layer norm that is used for normalization.
%
%
Lastly, we generate the output feature of cross-attention as
\begin{equation}
\begin{aligned}
\bar{\bm{ {F}}}^t_{r} = \bm{\hat{F}}^t_r + {\rm{Con}}{{\rm{v}}_{1 \times 1}}({\bm{V}_{r}}),
\end{aligned}
\end{equation}
where $\bm{{\hat{F}}}^t_r$ and $\bm{{\hat{F}}}^t_d$ are the inputs.  ${\bm{\bar{F}}_r^t\in \mathbb{R}^{\mathcal{W} \times \mathcal{H} \times \mathcal{D}}}$ and ${\bm{\bar{F}}_d^t\in \mathbb{R}^{\mathcal{W} \times \mathcal{H} \times \mathcal{D}}}$ are the respective outputs for cross-attention.
The cross-attention block is employed bidirectionally, facilitating information exchange in both directions:  RGB$\leftrightarrow$depth.

\subsection{Knowledge distillation and supervision} \label{sec:methods:Distillation Loss}
We train the model by minimizing the loss ${{\cal L}_{total}}$, which is composed of three components: ${{\cal L}_{k}}$ for knowledge distillation, ${{\cal L}_{p}}$ for pose estimation and ${{\cal L}_{o}}$ for optical flow.

\noindent\textbf{Multi-scale knowledge distillation for cross-modalities.}\label{sec:methods:Distillation}
The two-branch architecture is designed separately, resulting in features of different magnitudes\cite{crossdistillation,crossdistillation2}. This makes it difficult to aggregate and interact across modalities, affecting generalization capabilities.
To enhance the exchange between RGB and depth information, we optimize the network via multi-scale knowledge distillation at the feature level between ${\bm{\bar{F}}_r^t\in \mathbb{R}^{\mathcal{W} \times \mathcal{H} \times \mathcal{D}}}$ and ${\bm{\bar{F}}_d^t\in \mathbb{R}^{\mathcal{W} \times \mathcal{H} \times \mathcal{D}}}$.

The loss ${{\cal L}_{L2}}$ is formulated as the weighted sum of the L2 norm of the overall difference.  
\begin{equation}
\vspace{-1mm}
\begin{aligned}
{{\cal L}_{L2}} = \sum\limits_{c = 1}^{\mathcal{D}} {\sum\limits_{i = 1}^{\mathcal{H}} {\sum\limits_{j = 1}^{\mathcal{W}} {{{\left\| {\bm{\bar{F}}_r^t - \bm{\bar{F}}_d^t} \right\|}_2}} } }. 
\end{aligned}
\end{equation}
%
The loss ${{\cal L}_{s}}$ is designed to capture the spatial similarity between the mean feature maps of the RGB tensor ${\bm{\bar{F}}_r^t}$ and the depth tensor ${\bm{\bar{F}}_d^t}$ after applying a linear operation $\phi(\cdot)$. 
%
\begin{equation}
\begin{aligned}
    \mathcal{L}_{s} {=} \sum\limits_{c=1}^{\mathcal{D}} \left\| \frac{1}{\mathcal{HW}} \sum\limits_{i=1}^{\mathcal{H}} \sum\limits_{j=1}^{\mathcal{W}} \bm{\bar{F}}_r^t 
    {-} \phi{\left( \frac{1}{\mathcal{HW}} \sum\limits_{i=1}^{\mathcal{H}} \sum\limits_{j=1}^{\mathcal{W}} \bm{\bar{F}}_d^t \right)} \right\|.
\end{aligned}
\end{equation}
%
Loss ${{\cal L}_{c}}$ measures channel differences between target ${\bm{\bar{F}}_r^t}$ and ${\bm{\bar{F}}_d^t}$. It is calculated as the L2 norm of the discrepancy between the original feature maps and the convolved feature maps using a 1$\times$1 convolution, ${\rm{Con}}{{\rm{v}}_{1 \times 1}}(\cdot)$.
%
\begin{equation}
\begin{aligned}
{{\cal L}_{c}} = \sum\limits_{i = 1}^{\mathcal{H}} {\sum\limits_{j = 1}^{\mathcal{W}} {{{\left\| {\frac{1}{\mathcal{D}}\sum\limits_{c = 1}^{\mathcal{D}}{\bm{\bar{F}}_r^t} - {\rm{Con}}{{\rm{v}}_{1 \times 1}}\left(\frac{1}{\mathcal{D}}\sum\limits_{c = 1}^{\mathcal{D}}\bm{\bar{F}}_d^t\right)} \right\|}_2}} }.
\end{aligned}
\end{equation}
The overall knowledge distillation loss ${{\cal L}_{k}}$ is defined as
%
\begin{equation}
\begin{aligned}
{{\cal L}_{k}} = \alpha {{\cal L}_{L2}} + \beta {{\cal L}_{s}} + \delta {{\cal L}_{c}},
\end{aligned}
\end{equation}
where $\alpha$, $\beta$, and $\delta$ represent the weights of ${{\cal L}_{L2}}$, ${{\cal L}_{s}}$, and ${{\cal L}_{c}}$, respectively.

\noindent\textbf{Pose estimation loss.}\label{sec:methods:Distillation}
${{\cal L}_{p}}$ represents the loss between the estimated pose T and the ground truth G.
%
\begin{equation}
\begin{aligned} 
{{\cal L}_{p}} &= {\sum_{k = 1}^M \gamma^{M - k} {\left\| {{\rm{Lo}}{{\rm{g}}_{SE3}}({\rm{T}}_i^{ - 1} \cdot {{\rm{G}}_i})} \right\|} _2}.
\end{aligned}
\label{eq:loss}
\end{equation}

\noindent\textbf{Optical flow loss.} 
${{\cal L}_{o}}$ represents the loss between the estimated optical flow $(u_1', v_1')$ and the optical flow transformed by ground truth $(u_1, v_1)$. 
With increasing iteration $k$, the weight of each term increases exponentially with a base $\gamma$.
\begin{equation}
\begin{aligned}
{\cal L}_{{o}} &= \sum_{k = 1}^M \gamma^{M - k} \| (u_1', v_1') - (u_1, v_1) \|.
\end{aligned}
\label{eq:loss}
\end{equation}
Thus, the total loss ${{\cal L}_{total}}$ is calculated as:
\begin{equation}
\begin{aligned}
{{\cal L}_{total}} &= {{\cal L}_{k}} +{{\cal L}_{p}} + {{\cal L}_{o}}.
\end{aligned}
\label{eq:loss}
\vspace{-1mm}
\end{equation}

\subsection{Integrating GNSS-RTK and \fmname{} by global BA} \label{sec:methods:GNSS}

To integrate \fmname{} into the robot, we first fuse the position information between \fmname{} and GNSS-RTK in the robot by ulitizing the extended Kalman filter.
The state vector \( \mathit{{\rm{x}}_k} \) includes only the position coordinates, and the measurement vector \( \mathit{{\rm{z}}_k} \) consists of
$\left[ {\begin{array}{*{20}{c}}
{x_k^v}&{y_k^v}&{z_k^v}
\end{array}} \right]$ and $\left[ {\begin{array}{*{20}{c}}
{x_k^g}&{y_k^g}&{z_k^g}
\end{array}} \right]$, which are the position coordinates of visual SLAM and GNSS, respectively.
%
\begin{equation}
\renewcommand{\arraystretch}{1.1} 
\begin{aligned}
{{\rm{x}}_k} &= {\left[ {\begin{array}{*{20}{c}}
{{x_k}}&{{y_k}}&{{z_k}}
\end{array}} \right]^\top}, \\
{{\rm{z}}_k} &= {\left[ {\begin{array}{*{20}{c}}
{x_k^v}&{y_k^v}&{z_k^v}&{x_k^g}&{y_k^g}&{z_k^g}
\end{array}} \right]^\top}.
\end{aligned}
\end{equation}
The state prediction is performed as
\begin{equation}
\begin{aligned}
{{\rm{x}}_k} = f({\hat {\rm{x}}_{k - 1}},{u_k}),
{P_k} = F{\hat P_{k - 1}}{F^{\rm{T}}} + R,
\end{aligned}
\end{equation}
where ${u_k}$ represents the control vector, the error covariance matrix \( \mathit{P_k} \)  represents the estimated uncertainty of the state vector \( \mathit{{\rm{x}}_k} \), \( \mathit{F} \) represents the identity matrix, \( \mathit{R} \) is the process noise covariance matrix.

In the update step, the Kalman gain \( \mathit{K_k} \) is calculated using \( \mathit{P_k} \) and the Gaussian noise covariance matrix $Q$.
%
\begin{equation}
\begin{aligned}
{{{K}}_k} = {P_k}{{\rm{H}}^{\rm{T}}}{({\rm{H}}{P_k}{{\rm{H}}^{\rm{T}}} + {{Q}})^{ - 1}}.
\end{aligned}
\end{equation}
%
The state estimate ${\hat {\rm{x}}_k}$ is updated using the Kalman gain, ${\hat P_k}$ is updated to reflect the new estimation uncertainty.
%
\begin{equation}
\begin{aligned}
{\hat {\rm{x}}_k} = {{\rm{x}}_k} + {{\rm{K}}_k}({{\rm{z}}_k} - h({{\rm{x}}_k})),
{\hat P_k} = {P_k} - {{\rm{K}}_k}{\rm{H}}{P_k}.
\end{aligned}
\end{equation}

The fusion position coordinates ${\hat {\rm{x}}_k}$ are inserted into the pose matrix ${\hat {\rm{T}}_k}$. 
Then, global bundle adjustment (global BA) is performed by rebuilding the frame graph using the flow between all pairs of keyframes.

%% file: sections/exp_new.tex
\section{Experiments}\label{sec:exp}

We compare {\fmname} with state-of-the-art methods, including RESLAM\cite{RESLAM}, ORB-SLAM3\cite{ORB-SLAM3}, TartanVO\cite{Tartanvo}, and DROID-SLAM\cite{DROID_SLAM}. 
We conduct experiments using the TUM \cite{TUM} and TartanAir~\cite{TartanAir} datasets, and our dataset collected under indoor environments. 
Additionally, we validate our algorithm on a robot in outdoor environments.

\subsection{Experimental setup}

\subsubsection{Datasets}\label{sec:exp:datasets}

\begin{figure}[t]
\begin{center}
  \begin{tabular}{@{}c}
    \begin{overpic}[width=.432\columnwidth]{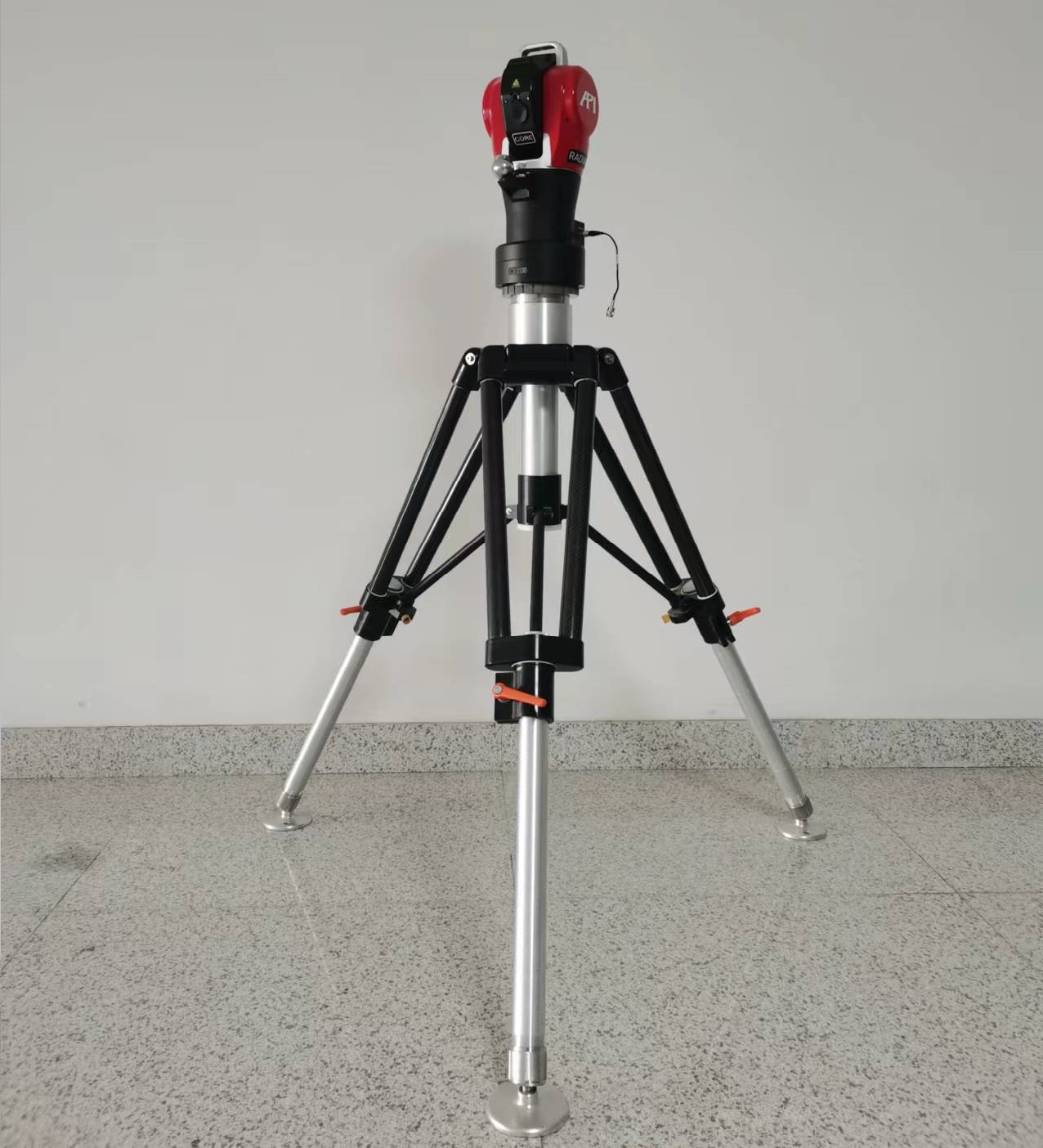}
    \put(0,-8){\color{black}\scriptsize\textbf{a) The laser tracker device}}
    \end{overpic}
    \begin{overpic}[width=.495\columnwidth]{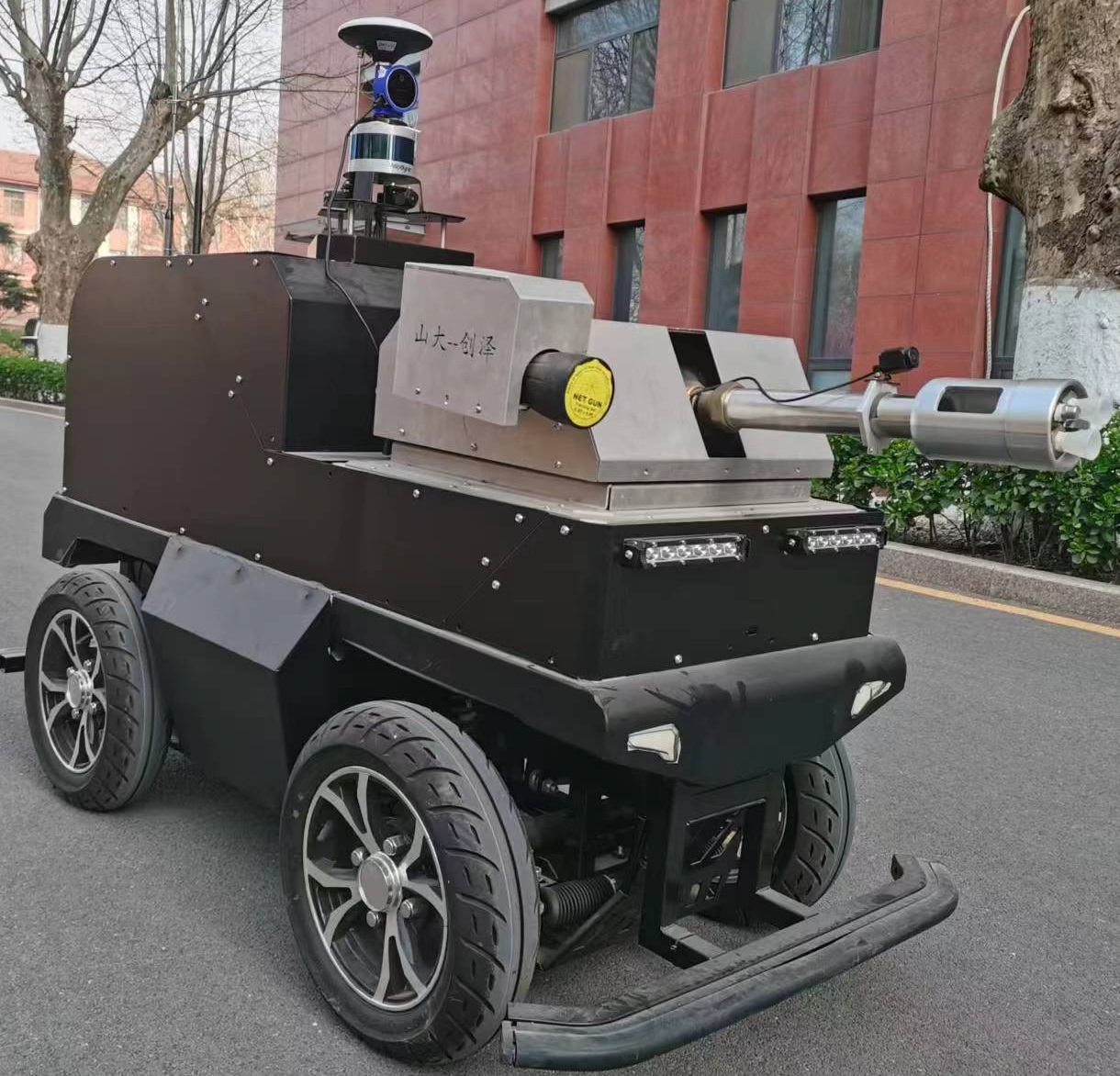}    
    \put(0,-8){\color{black}\scriptsize\textbf{b) The security robot}}
    \end{overpic}\\
  \end{tabular}
\end{center}
\caption{a) Front view of the laser tracker. b) Front view of our security robot.}
\label{fig:car}
\vspace{-4mm}
\end{figure}

The TUM dataset \cite{TUM} consists of indoor scenes captured with a handheld camera. The TartanAir dataset \cite{TartanAir} features diverse and challenging virtual environments. We divided the TartanAir dataset into training and test sets, with the test set containing 14 challenging sequences recorded under three distinct conditions: fast-moving dynamics, dimmed environments, and dark scenarios. For the fast-moving dynamics condition, we select sequences P09-P11 from Afactory and P05-P07 from Office. The dimmed environments are represented by sequences P01, P04, P06, and P08 from Endofworld. The dark scenarios are represented by sequences P08, P09, P10, and P13 from Afactorynight.
The remaining sequences are designated for the training set, ensuring that the test set does not overlap with the training set.
%
Our dataset is composed of two parts that are recorded with a RealSense L515 \cite{L515} camera under various lighting conditions in indoor environments with a resolution of 640$\times$480 pixels.
The first part is named Realtest-GT with ground truth in an indoor environment. 
The Realtest-GT dataset is composed of 6 sequences. 
Sequences 01-02, 03-04, and 05-06 are recorded under standard, light-changing, and dark conditions, respectively. 
Each sequence includes RGB-D images, along with the camera's trajectory recorded by a laser tracker\cite{tracker}, as shown in Fig.~\ref{fig:car} a). 
The trajectory recorded by the laser tracker is regarded as the ground truth for the datasets due to the high precision of the laser tracker.
The second part is named Realtest-Visual without ground truth. 
The Realtest-Visual dataset consists of three sequences with RGB-D images, which are recorded under standard, light-changing, and dark conditions, respectively. 
All the sequences follow a circular path back to the starting point.
The Realtest-GT and Realtest-Visual datasets enable us to quantitatively and qualitatively evaluate our method in real-world visually challenging scenes.
%

\subsubsection{Implementation details}\label{sec:exp:detail}

We implement \fmname{} in PyTorch. 
The network is trained on the TartanAir dataset with a batch size of 4 for 200K iterations, using 3 Nvidia 3090 GPUs for one week. 
For comparison methods, RGB-D information is used as input for RESLAM and ORB-SLAM3, while TartanVO relies solely on RGB data.
Both DROID-SLAM and our method are trained on the TartanAir dataset’s training set with identical initial parameters.
For evaluation metrics, we compare the absolute trajectory error between the estimated and ground-truth trajectories (ATE).

\vspace{-3mm}
\subsection{Comparisons on TUM and TartanAir}\label{sec:exp:comparisonsdataset}

\input{tables/table_sf_comparison}

\subsubsection{Quantitative results}\label{sec:exp:comparisons:Quantitative}
Tab.~\ref{tab:setting1_results} presents the ATE results of comparison algorithms under various challenging conditions in both the TUM and TartanAir datasets. We consider ATE values exceeding 100.00 cm or instances of tracking failure as outliers for the comparison algorithms. We also calculate the average ATE excluding these outliers. Fewer outliers indicate better robustness in localization under challenging environments, while a lower average ATE without outliers reflects higher localization accuracy.
Regarding the number of outliers, RESLAM, ORB-SLAM3, and TartanVO exhibit the highest counts. This is because these algorithms struggle to find the correct correspondences across frames. This illustrates that both classical SLAM algorithms and end-to-end learning-based SLAM algorithms have unstable localization performance in difficult environments. In contrast, our \fmname{} handles all challenging scenarios with zero outliers, demonstrating superior robustness compared to DROID-SLAM.
In terms of average ATE without outliers, RESLAM achieves an average ATE of 14.34 cm. TartanVO shows low accuracy across most evaluated sequences. ORB-SLAM3 delivers the second-best results with an average ATE of 2.77 cm, while our \fmname{} significantly outperforms it with an average ATE of 1.07 cm. Specifically, the ATE localization accuracy of our \fmname{} can achieve less than 1.00 cm, whereas the comparison algorithms frequently encounter tracking failures or significant localization errors, as seen in sequences \texttt{AfactoryP11} and \texttt{EndofworldP08}.
%
\begin{figure}[t]
\centering
\includegraphics[width=1\columnwidth]{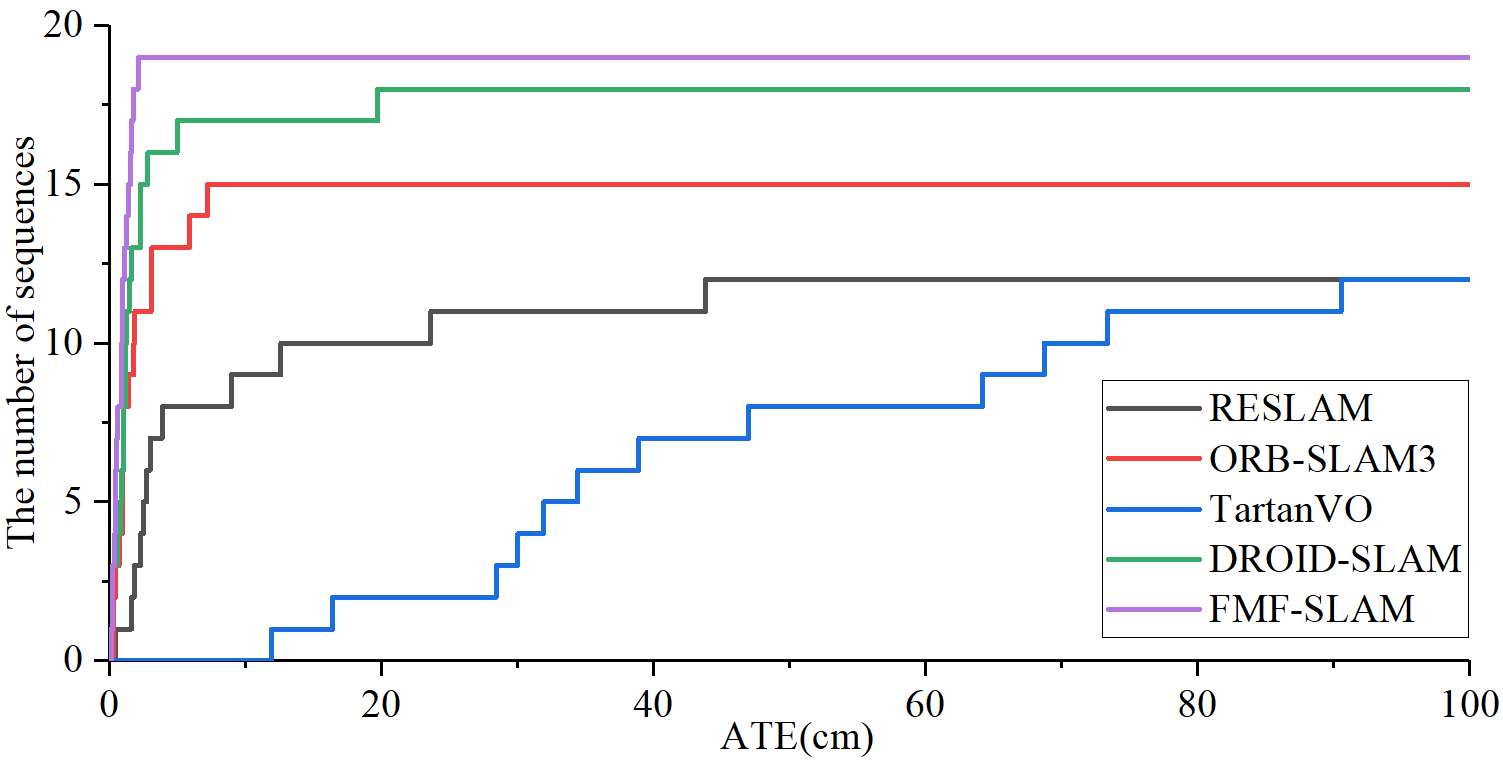}
\vspace{-6mm}
\caption{Relationship between the number of successfully tracked sequences (Tab.~\ref{tab:4}) and ATE for comparison methods. A larger area under the curve indicates better overall performance.}
\label{fig:method:plot1}
\vspace{-1mm}
\end{figure}
%

Fig.~\ref{fig:method:plot1} shows the relationship between the number of sequences successfully tracked and ATE for the comparison methods. 
\fmname{} achieves the largest area under the curve, demonstrating that it not only reduces outliers in various challenging environments but also achieves the most accurate localization with the lowest average ATE. 
%
The superior performance of \fmname{} is due to three key advantages. First, our method extracts features from both RGB and depth information, providing a rich and informative internal representation. 
Second, we utilize Fourier-based self-attention and cross-attention mechanisms, which effectively and efficiently highlight informative features across RGB and depth without downsampling. 
Third, multi-scale knowledge distillation between RGB and depth information helps maintain consistency and a shared understanding of cross-modality features, enabling a better the interaction and aggregation of the correlation between RGB and depth data.

\subsubsection{Qualitative results of optical flow estimation} 

Fig.~\ref{fig:optical_tartanair} shows optical flow visualizations of DROID-SLAM and our approach under fast-moving, dimmed, and dark conditions on the TartanAir dataset.
We observe that DROID-SLAM exhibits inadequate performance, as evidenced by the lack of clarity/details in the optical flow results for various objects. 
In contrast, \fmname{} consistently produces sharper edges, demonstrating that, compared to solely extracting features from RGB images, it can achieve more stable optical flow estimation results across various challenging environments.

\input{figures/figure_of_comparison_kitti}

\input{figures/dataset-tra}

\subsection{Comparisons in real-world indoor environments}\label{sec:exp:comparisonsindoor}

\subsubsection{Quantitative results on Realtest-GT}\label{sec:exp:comparisonsindoor:Realtest-GT}
\begin{figure}[t]
\centering
\includegraphics[width=.99\columnwidth]{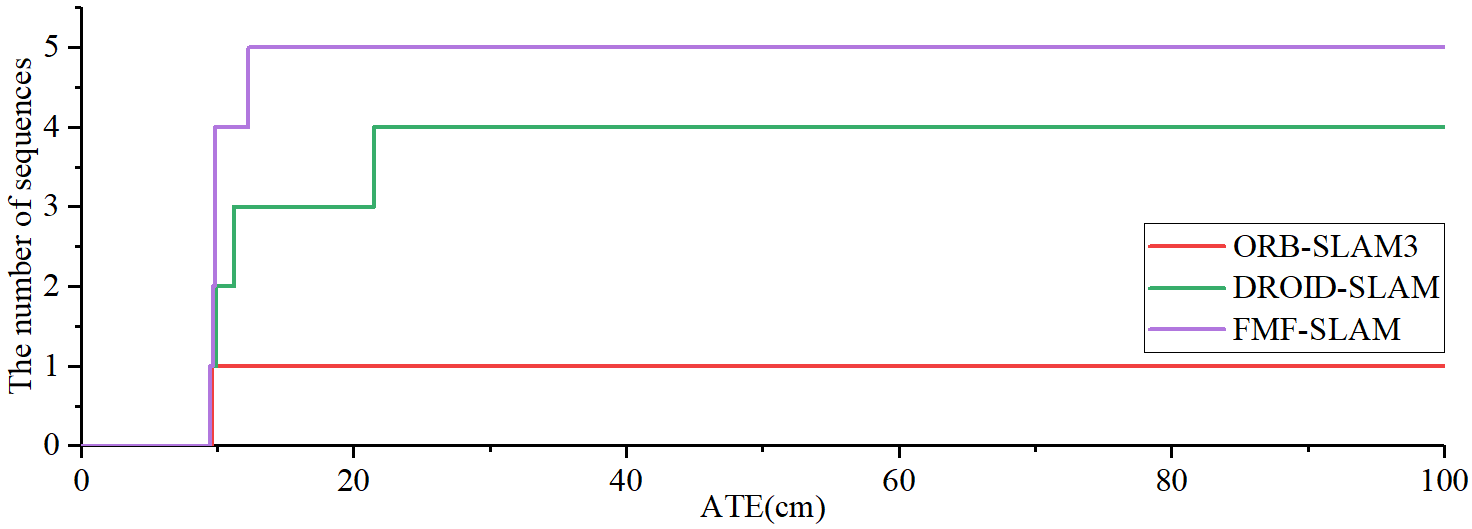}
\vspace{-6mm}
\caption{Relationship between successfully tracked sequences (Tab.~\ref{tab:setting2_results}) in the Realtest-GT dataset and ATE for comparison methods. A larger area under the curve indicates better overall performance.}
\label{fig:method:plot2}
\end{figure}

We compare our approach with RESLAM, ORB-SLAM3, TartanVO, and DROID-SLAM on the Realtest-GT dataset for ATE results.
Tab.~\ref{tab:setting2_results} reports the ATE results under standard, light changing, and dark conditions. RESLAM and ORB-SLAM3 fail to track due to the limited description based on edges or point features under challenging conditions, respectively. TartanVO and DROID-SLAM successfully track, but show inconsistent performance on sequences 03-06. 
This is because TartanVO and DROID-SLAM rely solely on features extracted from RGB images, which leads to unstable and weak feature descriptions in light-changing or dark environments. In contrast, our approach surpasses comparison methods with an average ATE of 11.03 cm and 16.84 cm under light-changing and dark conditions, respectively.
We also plot the relationship between the number of successfully tracked sequences in Realtest-GT and ATE in Fig.~\ref{fig:method:plot2} for ORB-SLAM3, DROID-SLAM, and our \fmname{}. 
Our \fmname{} achieves the largest area under the curve, demonstrating its ability to reduce outliers in challenging environments and achieve the most accurate localization with the lowest average ATE.
%
\input{tables/table_of_comparison}
%

\subsubsection{Qualitative results on Realtest-GT} 

We present the corresponding trajectories of sequence 02, sequence 04, and sequence 06 in the Realtest-GT data set in Fig.~\ref{fig:dataset-tra}, respectively.
It can be observed from Fig.~\ref{fig:dataset-tra} a) that the trajectories of all three methods are closely aligned with the ground truth under standard condition. 
While ORB-SLAM3 fails to track under light-changing and dark conditions, DROID-SLAM can track successfully but struggles to maintain consistent localization performance under challenging conditions. In contrast, the trajectories estimated by our method are close to the ground truth under both standard and challenging conditions.

\subsubsection{Qualitative results on Realtest-Visual} 
Fig.~\ref{fig:qualitative_res_traj} shows the reconstruction results in Realtest-Visual dataset. The three sequences record repeated scan regions under standard, light-changing, and dark conditions, respectively. We further evaluate the accuracy of our \fmname{}'s localization and mapping by observing the alignment in these repeated scan regions.
To highlight the alignment in overlapping regions, these parts are marked and displayed in a zoomed-in view within a blue bounding box. It is evident that our approach reconstructs the 3D map accurately and with precise alignment, proving the effectiveness of our \fmname{} under challenging conditions.
%
\input{figures/indoor_standard}
\vspace{-4mm}

\vspace{-1mm}
\subsection{Experiments on the security robot}\label{sec:exp:comparisons} 
We show the visualization results in large-scale outdoor environments by integrating our \fmname{} into a security robot.
Both sequences follow a circular path back to the starting point.
Due to the difficulty of obtaining stable depth information with the RealSense L515 in outdoor environments, we utilize ZoeDepth\cite{Zoedepth} to estimate depth from the RGB information and use both the RGB and estimated depth information as inputs for our \fmname{}.
The real-world demonstration underscores the robustness of \fmname{} in challenging  scenarios due to hardware constraints or environmental factors.
From the blue bounding boxes in Fig.~\ref{fig:outdoor3} a) and b), we can observe that our approach achieves excellent alignment in overlapping regions. 
We also report the corresponding distance between the endpoints of the estimated and GNSS-RTK trajectories in Tab.~\ref{tab:gnss}, noted as accumulation error. Our approach has low accumulation errors in large-scale environments.
These examples demonstrate that our approach can reliably perform localization and mapping in challenging outdoor environments. 
Moreover, our \fmname{} achieves 100.85 ms per frame, demonstrating that our approach can be deployed at real time.
\input{tables/table_outdoor}

%
\input{figures/out3}
\vspace{-3mm}

\subsection{Ablation study}

We conduct the ablation study on TartanAir to confirm the effectiveness of each module in Tab.~\ref{tab:performancebla}.
$\rm {ACC}_{1px}$ \cite{Fusionraft} measures the portion of errors that are within a threshold of one pixel.
$\rm AEPE_{2D}$ measures the average end-point error (EPE)~\cite{Fusionraft}, which is an average value of all the 2D flow errors.
$\rm Rot_{0.1^\circ}$ measures the portion of rotation errors that are within a threshold of 0.1 degrees.
$\rm Tra_{0.01m}$ measures the portion of translation errors that are within a threshold of 0.01 meters.

\input{tables/table_ablation_study}

We can observe that two-branch encoder for RGB information improves $\rm ACC_{1px}$ from $72.61{\rm{\% }}$ to $73.90{\rm{\% }}$ and reduces $\rm AEPE_{2D}$ from 3.10 to 2.66 (see Exp 1,2). 
Furthermore, the utilization of a two-branch encoder for both RGB and depth information consistently enhances the performance metrics of $\rm ACC_{1px}$ and $\rm AEPE_{2D}$. Applying the two-branch encoder to both RGB and depth yields superior results compared to applying it solely to RGB data (see Exp 2,3).
Overall, We can observe that all the components provide an incremental contribution to improving the quality of the optical flow and pose estimation results (see Exp 3,4,5). 
The best performance is
achieved when all the modules are activated, improving $\rm ACC_{1px}$ from $80.97{\rm{\% }}$ to $82.04{\rm{\% }}$ and reduce $\rm AEPE_{2D}$ from 1.75 to 1.51 with all the components together (see Exp 3,6). 

We conducted computational analysis with an Nvidia 3090 GPU (24GB) and an Intel i9-10900 CPU.
The total parameters and inference time of our encoder are 10.68 M and 9.91 ms, respectively.
The proposed \fmname{} achieves an average computation time of 77.76 ms per frame on TartanAir dataset.

%% file: tables/table_sf_comparison.tex
\begin{table*}[t]
    \tabcolsep 8pt
    \centering
    \caption{
Comparison of Absolute Trajectory Error-ATE (cm) on TUM and TartanAir datasets. And we selected three different challenging subsets from the TartanAir dataset for evaluation, each recorded under fast-moving, dimmed, and dark conditions.}
    \label{tab:setting1_results}
    \vspace{-.3cm}
    \label{tab:4}  
    \resizebox{0.7\linewidth}{!}{%
    \begin{tabular}{cl|ccccc}
        \toprule
        & \multirow{2}{*}{Sequence} & RESLAM & ORB-SLAM3  & TartanVO & DROID-SLAM& Our SLAM \\
        
        & &RGBD & RGBD&  RGB& RGB &RGBD\\

        \midrule
        \multirow{6}{*}{\rotatebox[origin=c]{90}{TUM}} & \texttt{\footnotesize fr1/desk} &2.73 & - &31.93 & \textbf{1.63} & 1.73 \\
        & \texttt{\footnotesize fr1/xyz} &2.31 & 0.97& 11.92& 1.04 & \textbf{0.96} \\
        & \texttt{\footnotesize fr2/xyz} &0.49 & 0.32& 28.43& \textbf{0.21} & 0.22 \\
        & \texttt{\footnotesize fr3/cabinet} &8.95 & -& 30.02& 1.25 & \textbf{1.10} \\
        & \texttt{\footnotesize fr3/office} &3.88 & \textbf{1.04}& 99.06& 1.18 & 1.28 \\
        & \texttt{\footnotesize fr3/far} &1.83 & 1.06& 16.41& 1.20 & \textbf{0.95} \\

        \midrule

        \multirow{6}{*}{\rotatebox[origin=c]{90}{Fast-moving}} & \texttt{\footnotesize AfactoryP09} &- & -& - & 19.73 & \textbf{3.89} \\
        & \texttt{\footnotesize AfactoryP10} &- & 3.11& - & 4.96 & \textbf{1.56} \\
        & \texttt{\footnotesize AfactoryP11} &43.81 &5.89& 64.16& 179.02 & \textbf{0.55} \\
        & \texttt{\footnotesize OfficeP05} &- & 0.46& 68.71& \textbf{0.21} & 0.22 \\
        & \texttt{\footnotesize OfficeP06} &- & 1.86& 108.56& 0.79 & \textbf{0.39} \\
        & \texttt{\footnotesize OfficeP07} &- & 0.32& 38.93 & \textbf{0.09} & 0.12 \\
        \midrule
        \multirow{4}{*}{\rotatebox[origin=c]{90}{Dimmed}}& \texttt{\footnotesize EndofworldP01} &1.62 & 1.42& 73.39& 1.47 & \textbf{0.44} \\
        & \texttt{\footnotesize EndofworldP04} &79.04 & 1.79& 239.42&2.82 & \textbf{1.63} \\
        & \texttt{\footnotesize EndofworldP06} &2.99 & 0.97& 90.53& 0.58 & \textbf{0.45} \\
        & \texttt{\footnotesize EndofworldP08} &23.63 & -& 143.63& 70.85 & \textbf{0.90} \\
        \midrule
        \multirow{4}{*}{\rotatebox[origin=c]{90}{Dark}}& \texttt{\footnotesize AfactorynightP08} &12.61 & 3.05& 46.96& \textbf{1.02} & 1.39 \\
        & \texttt{\footnotesize AfactorynightP09} &2.49 & \textbf{0.71} & 34.43& 2.27 & 0.90 \\
        & \texttt{\footnotesize AfactorynightP10} &- & 7.21 & -& 2.27 & \textbf{2.13} \\
        & \texttt{\footnotesize AfactorynightP13} &- & 14.07 & -& 0.86 & \textbf{0.61} \\
        \midrule
        & \multirow{1}{*}{Number of outliers} & 7 & 4  & 7 & 1& \textbf{0} \\
        \midrule 
        & \multirow{1}{*}{Avg. ATE w/o outliers} & 14.34 & 2.77  &48.84 & 6.02& \textbf{1.07} \\
        \bottomrule 
    \end{tabular}
    }
\end{table*}

%% file: figures/figure_of_comparison_kitti.tex
\begin{figure}[t]
\begin{center}
  \begin{tabular}{@{}c@{}c}
    \vspace*{2mm}
    \begin{overpic}[width=.23\linewidth]{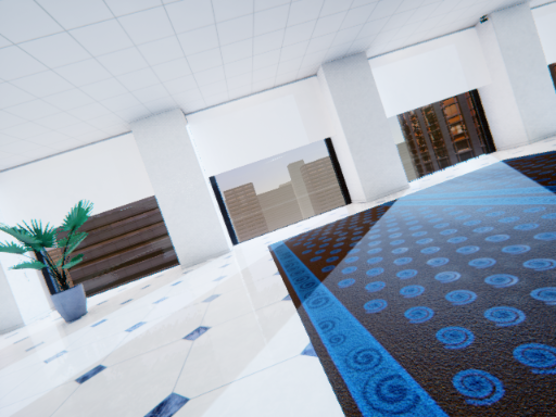}
     \put(35,80){\color{black}\scriptsize\textbf{RGB}}
     \put(0,-12){\color{black}\scriptsize\textbf{a) Fast-moving condition}}
    \end{overpic}
    \begin{overpic}[width=.23\linewidth]{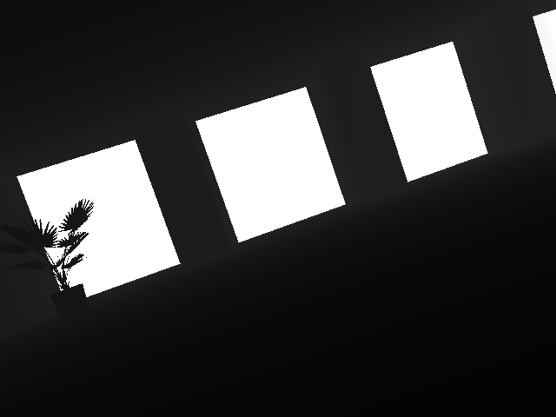}
    \put(35,80){\color{black}\scriptsize\textbf{depth}}
    \end{overpic}
    \begin{overpic}[width=.23\linewidth]{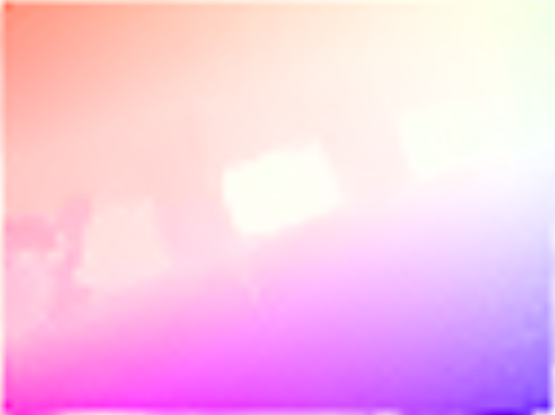}
    \put(15,80){\color{black}\scriptsize\textbf{DROID-SLAM}}
    \end{overpic}
    \begin{overpic}[width=.23\linewidth]{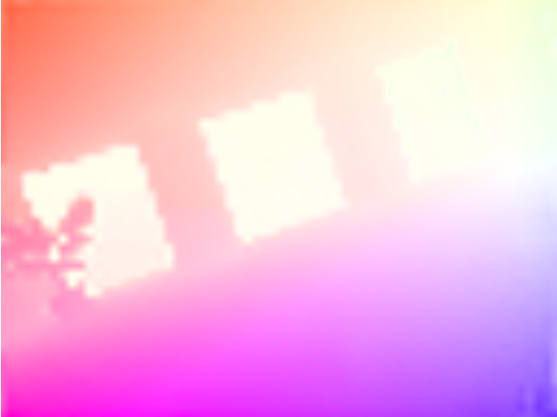}
    \put(18,80){\color{black}\scriptsize\textbf{{\fmname}}}
    \end{overpic}\\
    \vspace*{2mm}
    \begin{overpic}[width=.23\linewidth]{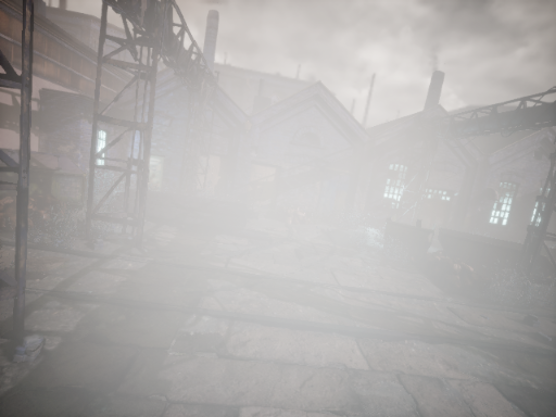}
    \put(0,-12){\color{black}\scriptsize\textbf{b) Dimmed condition}}
    \end{overpic}
    \begin{overpic}[width=.23\linewidth]{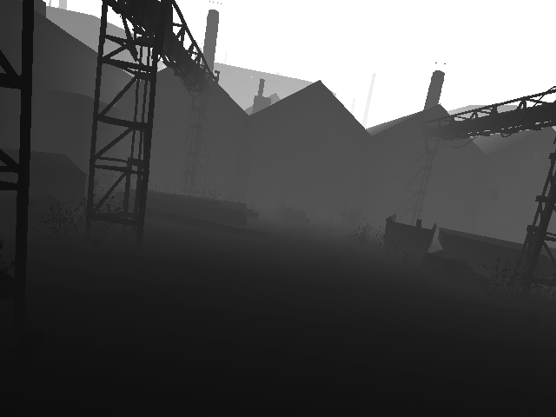}
    \end{overpic}
    \begin{overpic}[width=.23\linewidth]{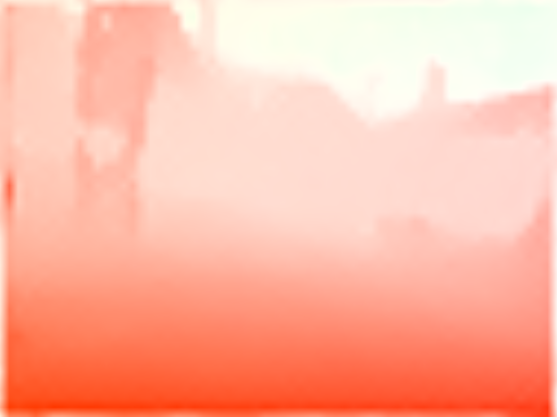}
    \end{overpic}
    \begin{overpic}[width=.23\linewidth]{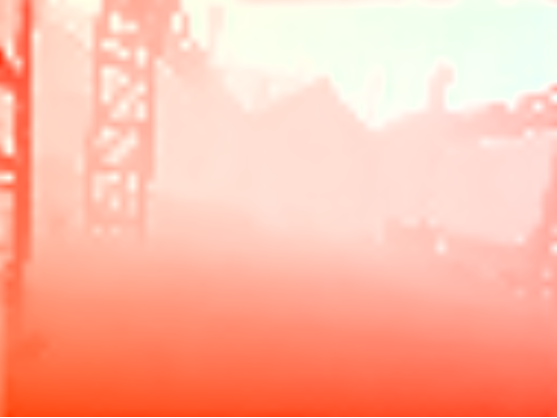}
    \end{overpic}\\
    \vspace*{2mm}
    \begin{overpic}[width=.23\linewidth]{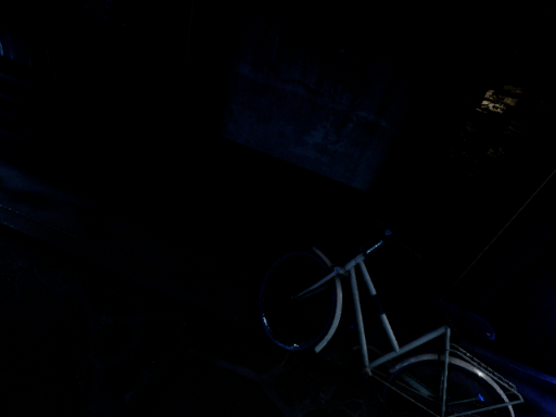}
    \put(0,-12){\color{black}\scriptsize\textbf{c) Dark condition}}
    \end{overpic}
    \begin{overpic}[width=.23\linewidth]{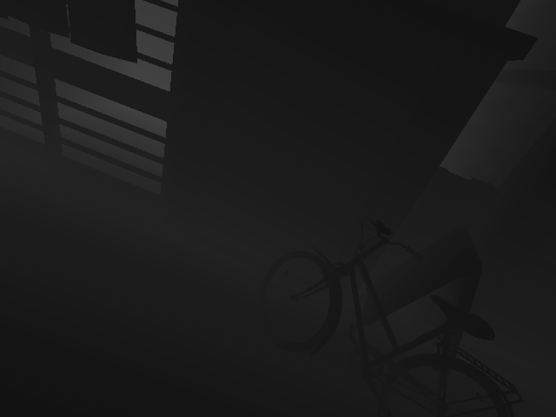}
    \end{overpic}
    \begin{overpic}[width=.23\linewidth]{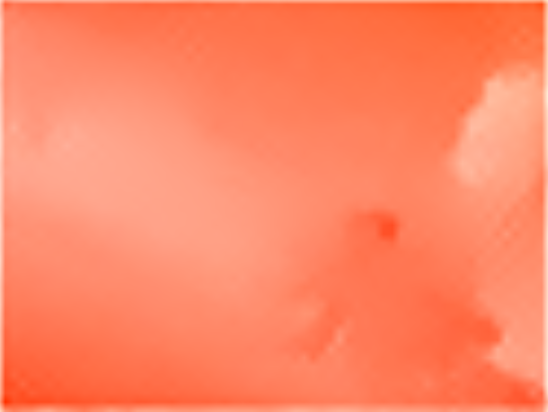}
    \end{overpic}
    \begin{overpic}[width=.23\linewidth]{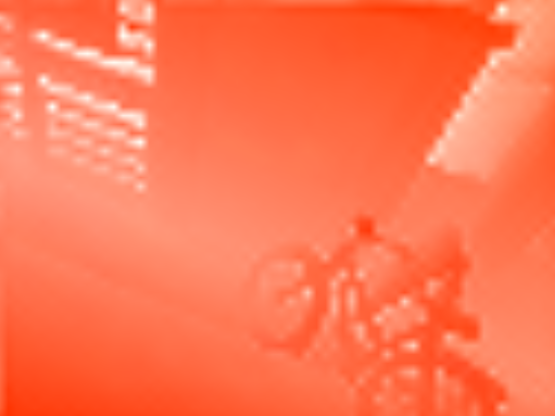}
    \end{overpic}\\
  \end{tabular}
\end{center}
\vspace{-2mm}
\caption{
Examples of optical flow estimation under various environments on TartanAir dataset. The clearer the edges, the higher the optical flow quality. The blurrier the boundaries, the poorer the optical flow quality.
}
\label{fig:optical_tartanair}
\vspace{-3mm}
\end{figure}
\vspace{-4mm}

%% file: figures/dataset-tra.tex
\begin{figure}[t]
\begin{center}
  \begin{tabular}{@{}c@{\,\,\,\,\,}c@{}c@{}c}

  \adjustbox{raise=0.2cm}{\begin{overpic}[width=.5\columnwidth]{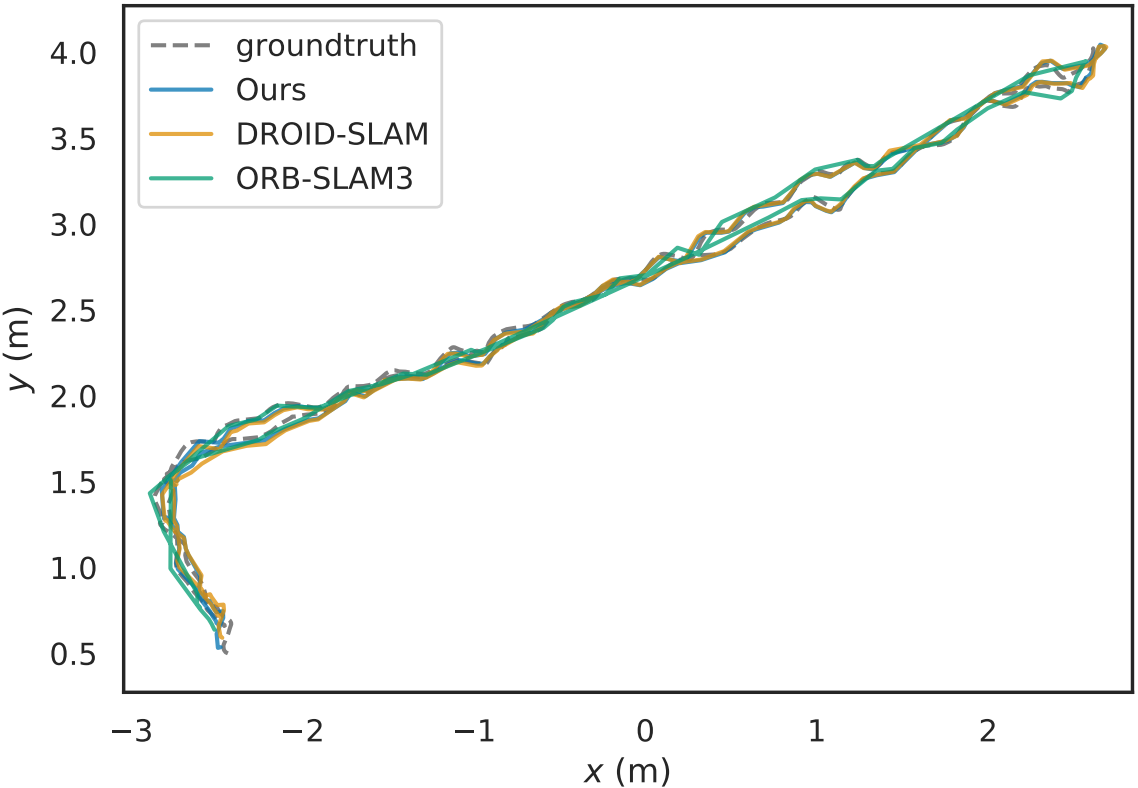}
      \put(9,17){
            \begin{tikzpicture}
            \draw[blue, very thick, opacity=0.65] (0,0) rectangle (1.28,.97);
            \end{tikzpicture}
        }
        \put(-5,18){\color{black}\scriptsize\rotatebox{90}{\textbf{a) Sequence 02}}}
    \end{overpic}}
    \adjustbox{raise=.55cm}{\begin{overpic}[width=.43\columnwidth, height=.3\columnwidth]{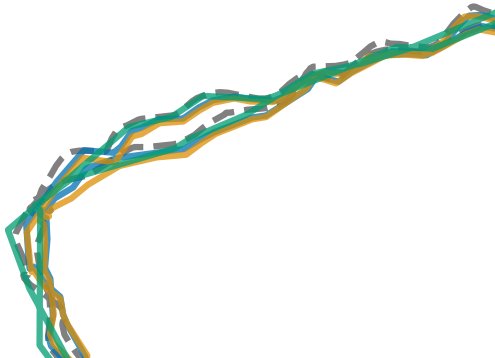}
        \put(-3,0){
                \begin{tikzpicture}
                \draw[blue, very thick, opacity=0.65] (0,0) rectangle (3.8,2.66);
                \end{tikzpicture}
                }
        \put(18,-7){\color{black}\scriptsize\textbf{zoomed-in bounding box}}
    \end{overpic}}\\

\adjustbox{raise=0.2cm}{\begin{overpic}[width=.5\columnwidth]{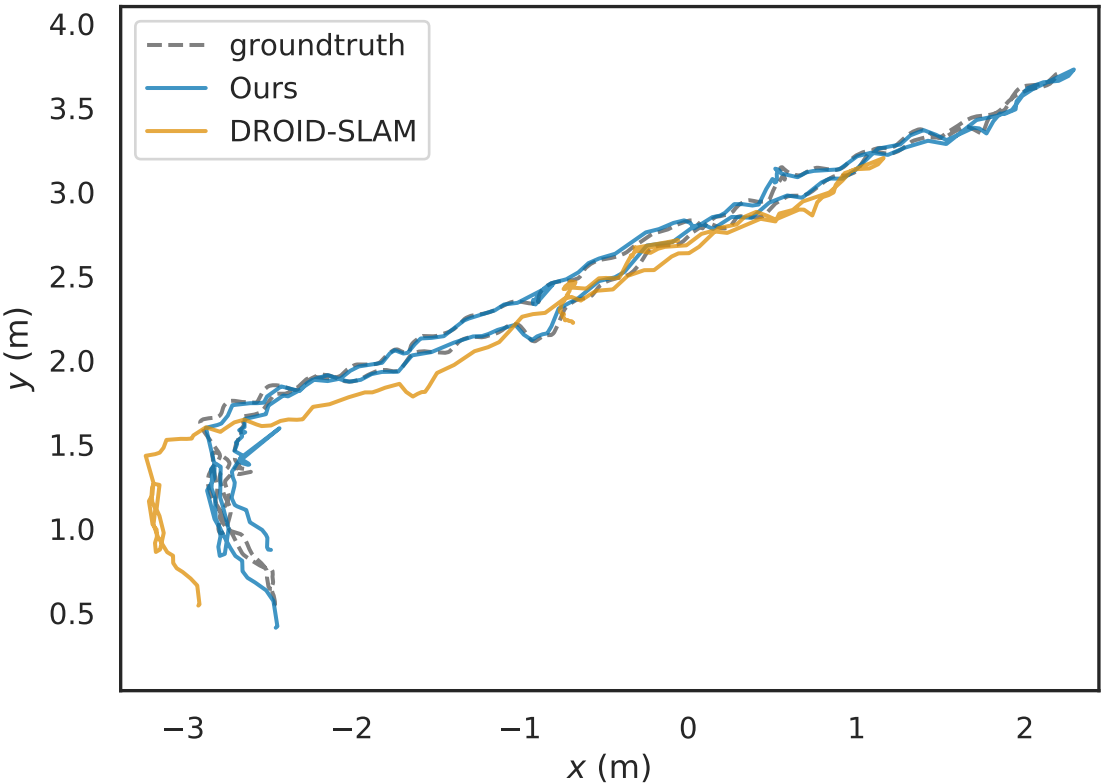}
      \put(32,33){
            \begin{tikzpicture}
            \draw[blue, very thick, opacity=0.65] (0,0) rectangle (1.28,.85);
            \end{tikzpicture}
        }
        \put(-5,18){\color{black}\scriptsize\rotatebox{90}{\textbf{b) Sequence 04}}}
    \end{overpic}}
    \adjustbox{raise=.55cm}{\begin{overpic}[width=.43\columnwidth, height=.3\columnwidth]{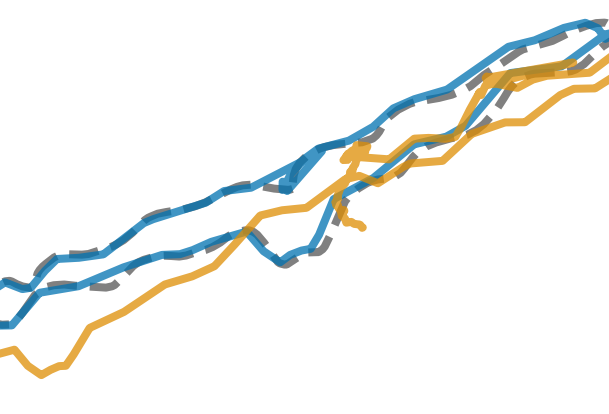}
        \put(-3,0){
                \begin{tikzpicture}
                \draw[blue, very thick, opacity=0.65] (0,0) rectangle (3.8,2.75);
                \end{tikzpicture}
                }
        \put(18,-7){\color{black}\scriptsize\textbf{zoomed-in bounding box}}
    \end{overpic}}\\
    \adjustbox{raise=0.2cm}{\begin{overpic}[width=.5\columnwidth]{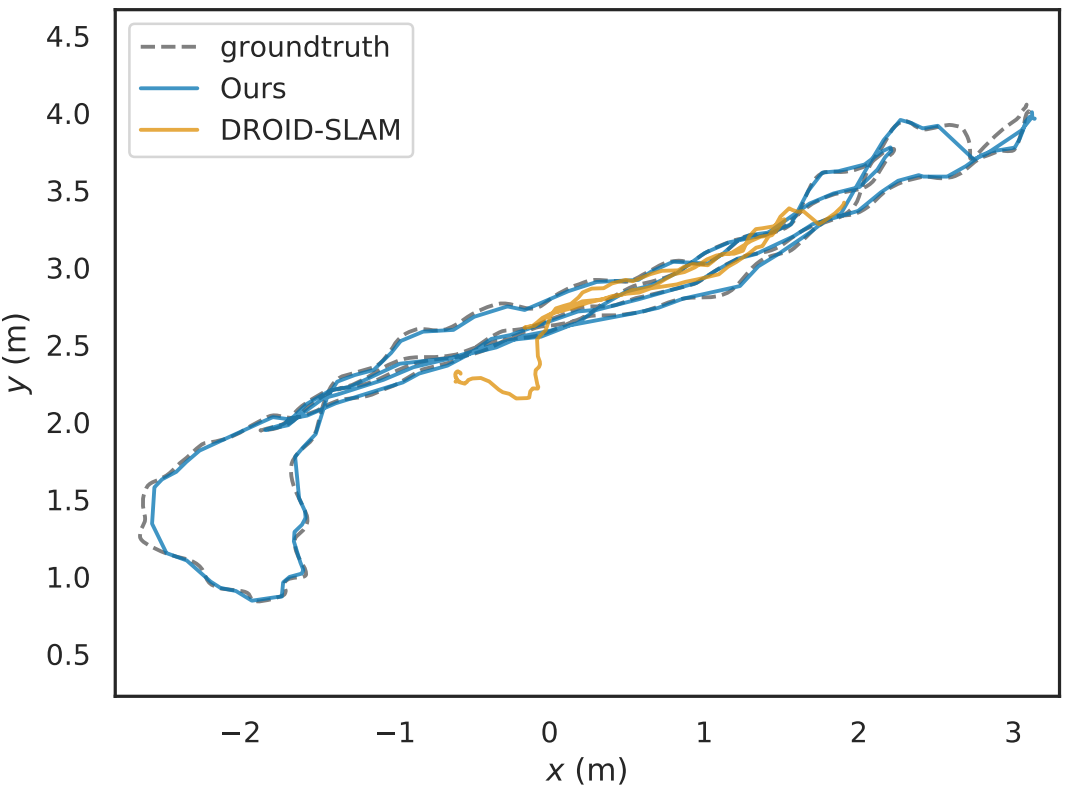}
      \put(39,34){
            \begin{tikzpicture}
            \draw[blue, very thick, opacity=0.65] (0,0) rectangle (1.40,.96);
            \end{tikzpicture}
        }
        \put(-5,18){\color{black}\scriptsize\rotatebox{90}{\textbf{c) Sequence 06}}}
    \end{overpic}}
    \adjustbox{raise=.55cm}{\begin{overpic}[width=.43\columnwidth, height=.3\columnwidth]{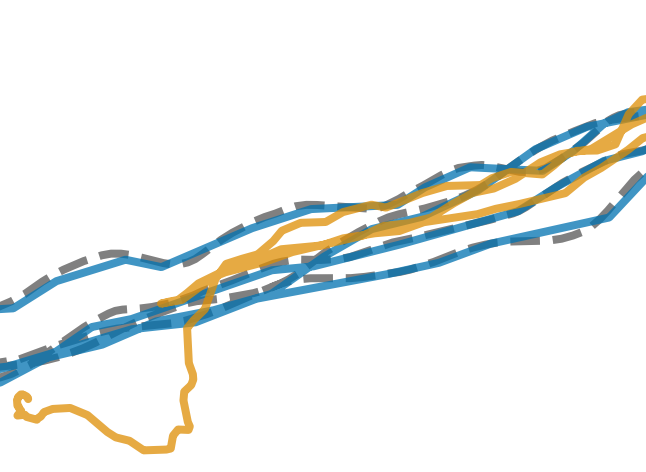}
        \put(-3,0){
                \begin{tikzpicture}
                \draw[blue, very thick, opacity=0.65] (0,0) rectangle (3.8,2.85);
                \end{tikzpicture}
                }
        \put(18,-7){\color{black}\scriptsize\textbf{zoomed-in bounding box}}
    \end{overpic}}
  \end{tabular}
\end{center}
\vspace{-5mm}
\caption{Comparison between the groundtruth and estimated trajectories for state-of-the-art SLAM
systems on Realtest-GT dataset.}
\label{fig:dataset-tra}
\vspace{-4mm}
\end{figure}

%% file: tables/table_of_comparison.tex
\begin{table}[t]
    \tabcolsep 4pt
    \centering
    \caption{
    Absolute Trajectory Error-ATE (cm) comparison on Realtest-GT dataset.}
    \label{tab:setting2_results}
    \vspace{-.3cm}
    \resizebox{\linewidth}{!}{%
    \begin{tabular}{c|cccccc}
        \toprule
         \multirow{2}{*}{Seq.} & RESLAM & ORB-SLAM3 & TartanVO  & DROID-SLAM& Our SLAM \\
         &RGBD & RGBD&  RGB &RGBD&RGBD\\
        \midrule
         \texttt{\footnotesize 01} &53.21 & 10.53& 49.14 & 9.43 & \textbf{9.42} \\
         \texttt{\footnotesize 02} &81.56 & 9.62&77.34 & 9.90 & \textbf{9.65} \\
        \midrule
          \texttt{\footnotesize 03} &- & -&45.22 & 11.19 & \textbf{9.79} \\
         \texttt{\footnotesize 04} &- & -&104.55 & 123.06 & \textbf{12.26} \\
        \midrule
         \texttt{\footnotesize 05} &- & -& 64.00&  21.45 & \textbf{9.82} \\
         \texttt{\footnotesize 06} & - & -& 51.75 &  177.88 & \textbf{23.86} \\
        \bottomrule 
    \end{tabular}
    }
    \vspace{-3mm}
\end{table}

%% file: figures/indoor_standard.tex
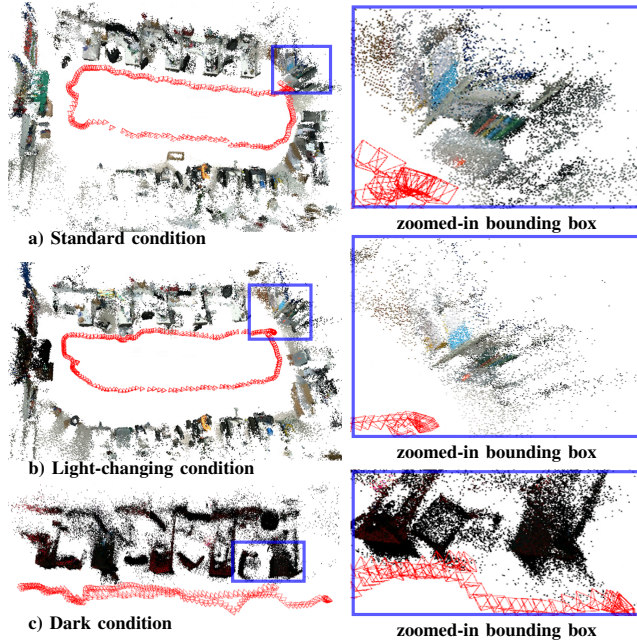
\begin{figure}[t]
\begin{center}
  \begin{tabular}{@{}c@{\,\,\,\,\,}c@{}c@{}c}
    \adjustbox{raise=0cm}{\begin{overpic}[width=.50\columnwidth]{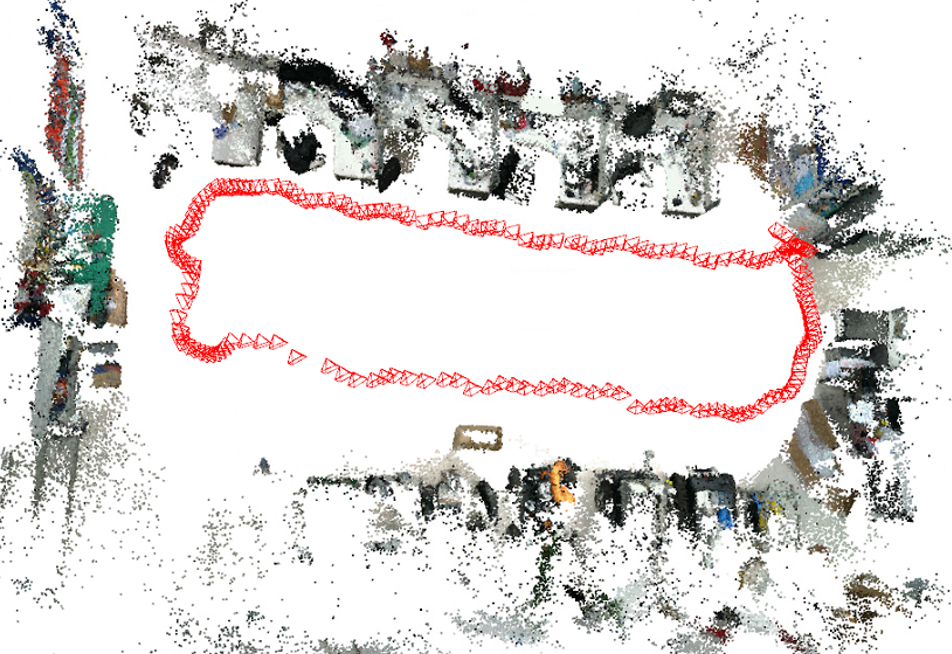}
      \put(76,41){
            \begin{tikzpicture}
            \draw[blue, very thick, opacity=0.65] (0,0) rectangle (.77,.62);
            \end{tikzpicture}
        }
        \put(6,-4){\color{black}\scriptsize\textbf{a) Standard condition}}
    \end{overpic}}
    \adjustbox{raise=0.3cm}{\begin{overpic}[width=.43\columnwidth, height=.3\columnwidth]{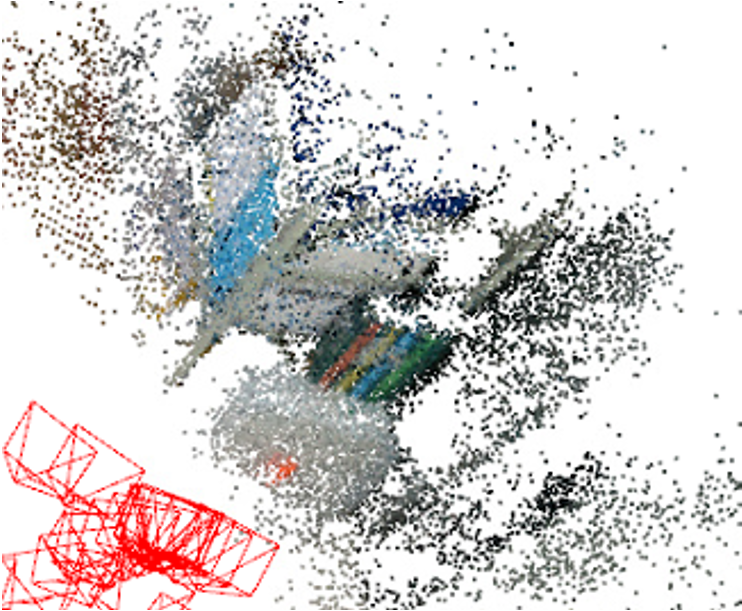}
        \put(-3,0){
                \begin{tikzpicture}
                \draw[blue, very thick, opacity=0.65] (0,0) rectangle (3.8,2.66);
                \end{tikzpicture}
                }
        \put(16,-7){\color{black}\scriptsize\textbf{zoomed-in bounding box}}
    \end{overpic}}\\
    \adjustbox{raise=0cm}{\begin{overpic}[width=.50\columnwidth]{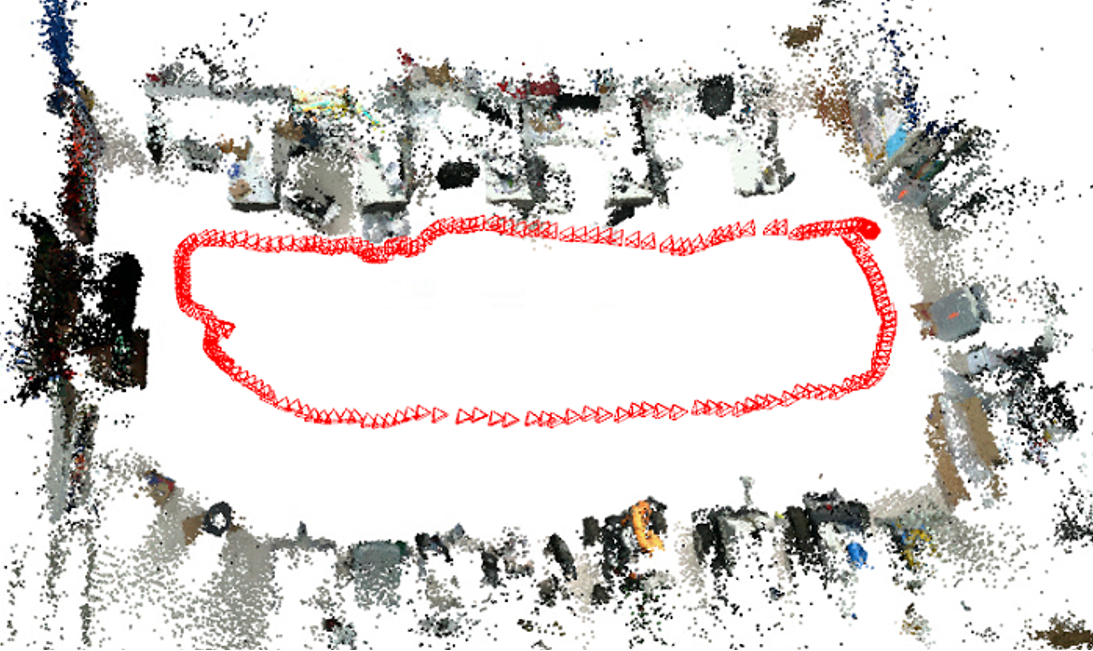}
      \put(69,36){
            \begin{tikzpicture}
            \draw[blue, very thick, opacity=0.65] (0,0) rectangle (.82,.72);
            \end{tikzpicture}
        }
        \put(6,-4){\color{black}\scriptsize\textbf{b) Light-changing condition}}
    \end{overpic}}
    \adjustbox{raise=0.3cm}{\begin{overpic}[width=.43\columnwidth, height=.3\columnwidth]{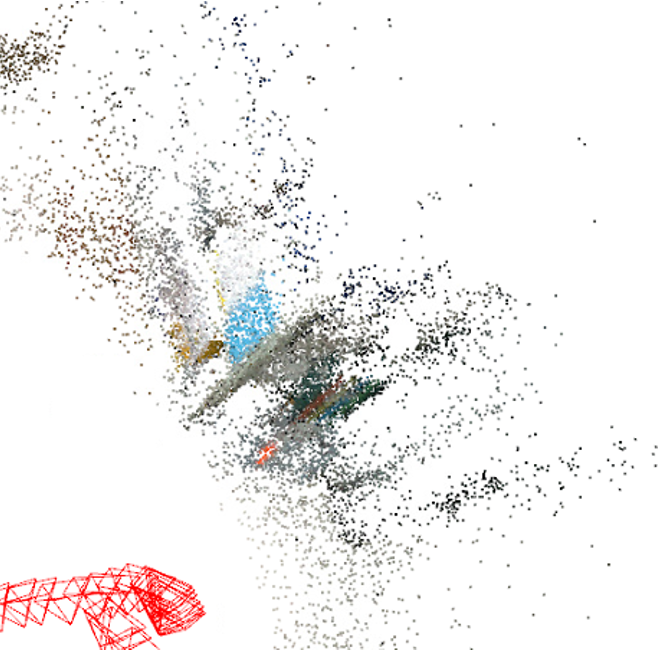}
        \put(-3,0){
                \begin{tikzpicture}
                \draw[blue, very thick, opacity=0.65] (0,0) rectangle (3.8,2.66);
                \end{tikzpicture}
                }
        \put(16,-7){\color{black}\scriptsize\textbf{zoomed-in bounding box}}
    \end{overpic}}\\
    \adjustbox{raise=0.15cm}{\begin{overpic}[width=.50\columnwidth, height=.2\columnwidth]{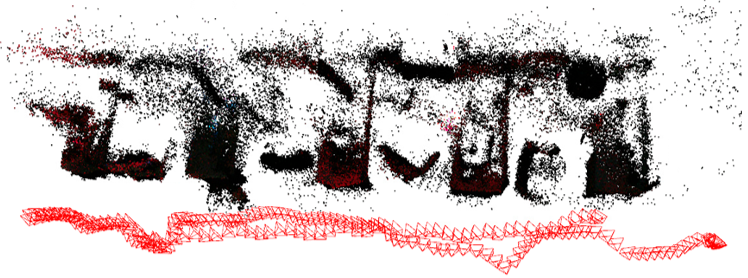}
      \put(64,10){
            \begin{tikzpicture}
            \draw[blue, very thick, opacity=0.65] (0,0) rectangle (.95,.52);
            \end{tikzpicture}
        }
        \put(6,-4){\color{black}\scriptsize\textbf{c) Dark condition}}
    \end{overpic}}
    \adjustbox{raise=0.15cm}{\begin{overpic}[width=.43\columnwidth, height=.22\columnwidth]{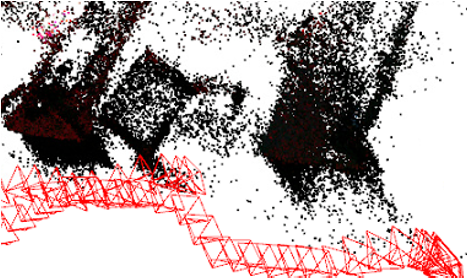}
        \put(-3,0){
                \begin{tikzpicture}
                \draw[blue, very thick, opacity=0.65] (0,0) rectangle (3.8,1.90);
                \end{tikzpicture}
                }
        \put(16,-7){\color{black}\scriptsize\textbf{zoomed-in bounding box}}
    \end{overpic}}
  \end{tabular}
\end{center}
\vspace{-3mm}
\caption{3D reconstruction of our \fmname{} under standard, light-changing, and dark conditions, respectively.}
\label{fig:qualitative_res_traj}
\vspace{-3mm}
\end{figure}

%% file: tables/table_outdoor.tex
\renewcommand{\arraystretch}{.9}
\begin{table}[t]
\small
    \centering
    \caption{Error accumulation of \fmname{} for large-scale cases.}
    \vspace{-2mm}
    \label{tab:gnss}  
    {%
    \begin{tabular}{lccc}
        \toprule
        Sequence & Distance [m] &Accumulation error [m]& Proportion \\
        \midrule
        \texttt{\footnotesize 01} & 147.45 & 0.09 &0.06 \%\\ 
        \texttt{\footnotesize 02} & 391.82& 0.26 &0.07 \%\\
        \bottomrule 
    \end{tabular}
    }
    \vspace{-2mm}
\end{table}

%% file: figures/out3.tex
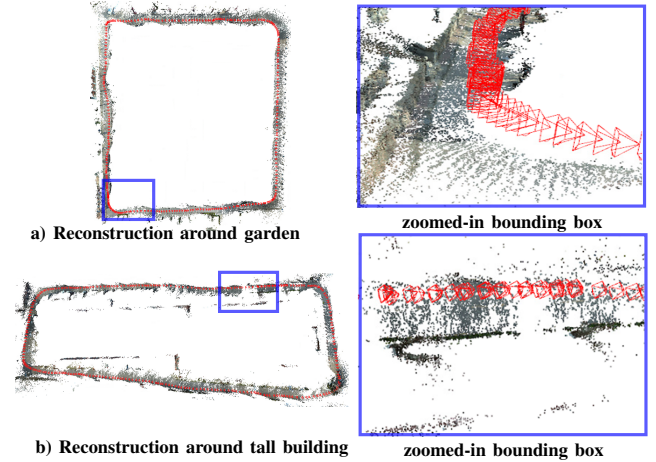
\begin{figure}[t]
\begin{center}
  \begin{tabular}{@{}c@{\,\,\,\,\,}c@{}c@{}c}
    \hspace{0.85cm}
    \adjustbox{raise=0cm}{\begin{overpic}[width=.3\columnwidth]{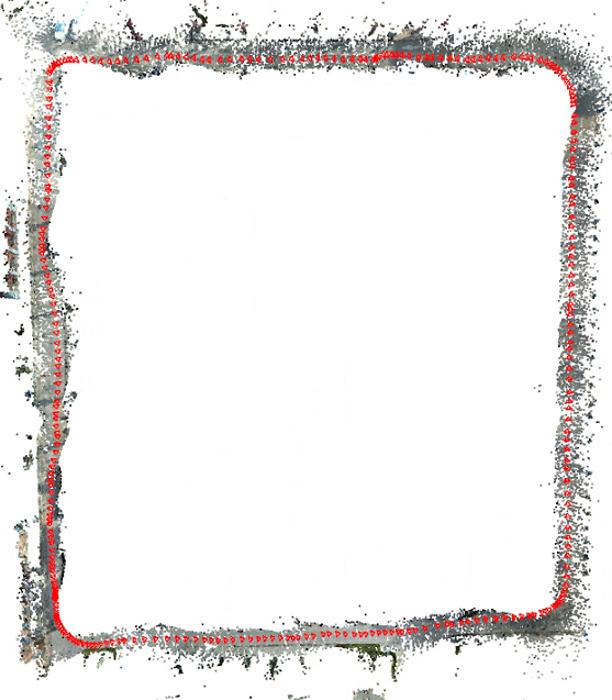}
      \put(1,4){
            \begin{tikzpicture}
            \draw[blue, very thick, opacity=0.65] (0,0) rectangle (.67,.52);
            \end{tikzpicture}
        }
        \put(-26,-4){\color{black}\scriptsize\textbf{a) Reconstruction around garden}}
    \end{overpic}}
    \hspace{0.65cm} 
    \adjustbox{raise=0.3cm}{\begin{overpic}[width=.43\columnwidth, height=.3\columnwidth]{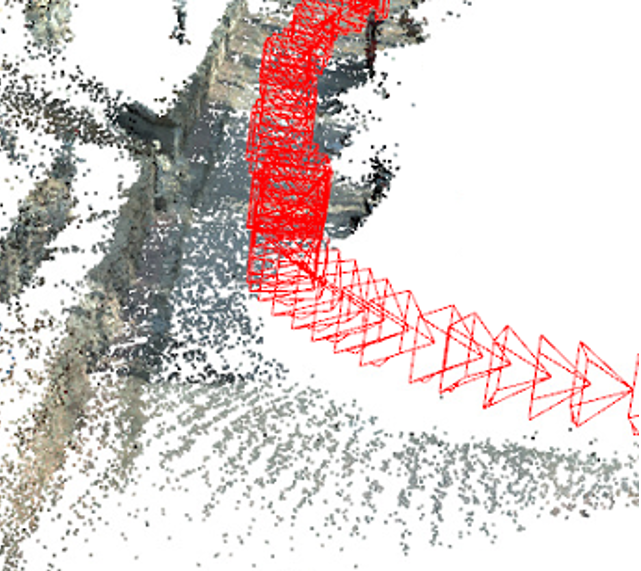}
        \put(-3,0){
                \begin{tikzpicture}
                \draw[blue, very thick, opacity=0.65] (0,0) rectangle (3.8,2.66);
                \end{tikzpicture}
                }
        \put(16,-7){\color{black}\scriptsize\textbf{zoomed-in bounding box}}
    \end{overpic}}\\
    \adjustbox{raise=0.4cm}{\begin{overpic}[width=.5\columnwidth]{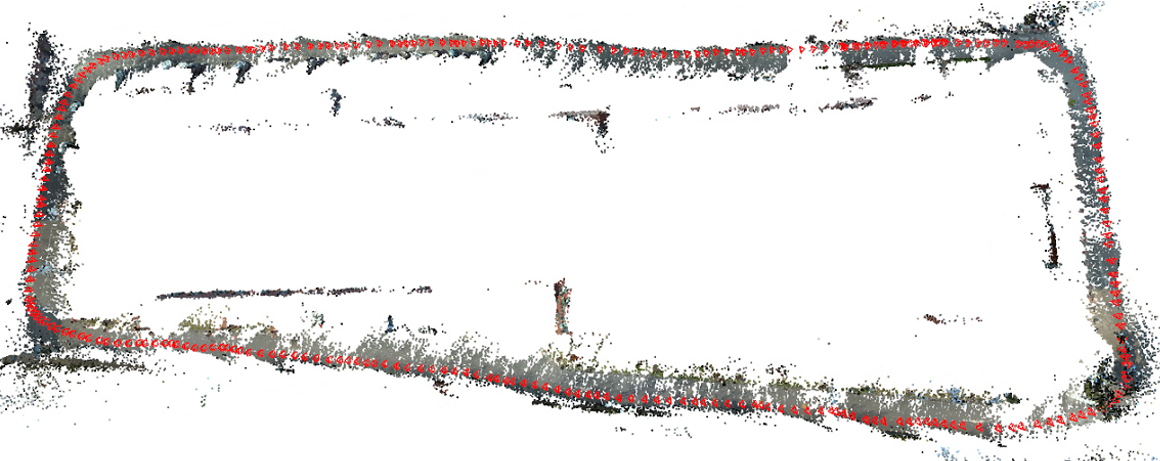}
      \put(58,28){
            \begin{tikzpicture}
            \draw[blue, very thick, opacity=0.65] (0,0) rectangle (.77,.52);
            \end{tikzpicture}
        }
        \put(6,-14){\color{black}\scriptsize\textbf{b) Reconstruction around tall building}}
    \end{overpic}}
    \adjustbox{raise=0.0cm}{\begin{overpic}[width=.43\columnwidth, height=.3\columnwidth]{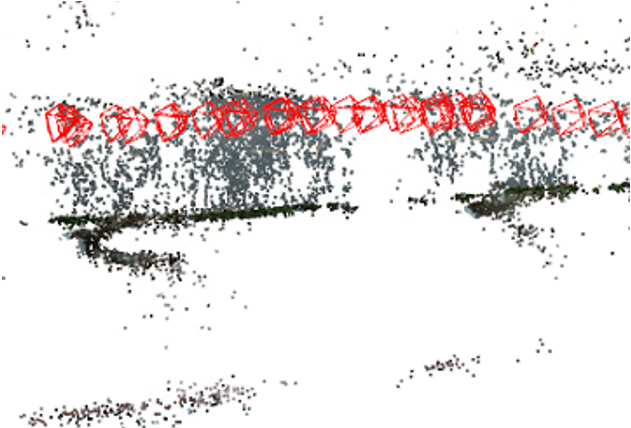}
        \put(-3,0){
                \begin{tikzpicture}
                \draw[blue, very thick, opacity=0.65] (0,0) rectangle (3.8,2.66);
                \end{tikzpicture}
                }
        \put(16,-7){\color{black}\scriptsize\textbf{zoomed-in bounding box}}
    \end{overpic}}\\
  \end{tabular}
\end{center}
\caption{3D reconstruction of our \fmname{} around
garden and tall building in outdoor environments.}
\vspace{-0.45cm}
\label{fig:outdoor3}
\end{figure}

%% file: tables/table_ablation_study.tex
\renewcommand{\arraystretch}{.9}
\begin{table}[t]
    \centering
    \tabcolsep 3pt
    \caption{
    Ablation study of our multimodal fusion encoder on the TartanAir dataset. R and D denote RGB and depth, respectively. FA is Fourier attention, Dis is distillation, and FMF is the feature encoder combining FA and Dis.
    }
    \vspace{-3mm}
    \label{tab:performancebla}  
    \resizebox{\linewidth}{!}{%
    \begin{tabular}{lccccc}
        \toprule
        Exp & Model & $\rm ACC_{1px}$ [\%] & $\rm AEPE_{2D}$ (pixel) & $\rm Rot_{0.1^\circ}$ [\%] & $\rm Tra_{0.01m}$ [\%] \\
        \midrule
        1&R & 72.61 & 3.10& 86.02&97.00\\
        2&R+R & 73.90 & 2.66& 86.48&97.95\\
        3&R+D & 80.97 &1.75 & 89.27&98.45\\
        4&R+D+FA & 81.68 &1.59 & 90.21&98.34\\
        5&R+D+Dis & 81.33& 1.65& 90.36&98.43 \\
        6&R+D+FMF & 82.04& 1.51& 90.73&98.45 \\
        \bottomrule 
    \end{tabular}
    }
    \vspace{-3mm}
\end{table}

%% file: sections/conclusions.tex
\section{Conclusions}\label{sec:conclusion}

\vspace{-.1cm}

In this paper, we propose \fmname{} that performs localization and mapping for indoor and outdoor environments.
To handle noisy, light-changing, and dark environments, as well as maintain real-time performance for SLAM, we introduced a novel Fourier attention mechanism to efficiently obtain informative features across modalities. We further introduced a multi-scale knowledge distillation to facilitate effective cross-modal feature fusion.
The experimental results, conducted on TUM, TartanAir, and our real-world datasets with various challenging conditions, underscore that our \fmname{} outperforms existing state-of-the-art learning-based SLAM systems in both standard and challenging conditions.
Finally, we integrated our \fmname{} onto a security robot by fusing GNSS-RTK based on the constraints of global BA, achieving reliable localization and mapping in real time.
